  \documentclass{EngC}

  \usepackage[rightcaption,raggedright]{sidecap}
  \usepackage{framed}         
  \usepackage{soul}           

  \usepackage{rotating}
  \usepackage{floatpag}
  \rotfloatpagestyle{empty}
  \usepackage{amsmath}
  \usepackage{amsthm}
  \usepackage{graphicx}

  \usepackage{amssymb}
  \usepackage{bm}
  \usepackage{amsfonts}
  \usepackage{tabularx}
  \usepackage{multirow}
  \usepackage{wrapfig,subfigure}
  \usepackage{stfloats}
  \usepackage{algorithm}
  \usepackage{fancyhdr, epsfig, epsf, subfigure}
  \usepackage{dsfont}
  \usepackage[noend]{algpseudocode}
  \usepackage{booktabs}
  \usepackage{rotating,booktabs,multirow}
  \usepackage{makecell}
  
\def\({\left(}
\def\){\right)}

\def\r{\right}

  \usepackage{multind}
  \makeindex{authors}
  \makeindex{subject}

  \theoremstyle{plain}

  \theoremstyle{definition}

  \theoremstyle{remark}
  \newtheorem{remark}{Remark}[chapter]

\begin{document}

\alphafootnotes

  \chapterauthor{Hyowoon Seo\footnotemark,\
  Jihong Park\footnotemark,\
  Seungeun Oh\footnotemark,\\
  Mehdi Bennis\footnotemark,\
  and Seong-Lyun Kim$^c$
    }

  \chapter{Federated Knowledge Distillation}

  \footnotetext[1]{H. Seo was with the Department of Electrical and Computer Engineering, Seoul National University, and is now with the Centre for Wireless Communications, University of Oulu, Oulu 90014, Finland (email: hyowoon.seo@oulu.fi).
  }
  \footnotetext[2]{J. Park is with the School of Information Technology, Deakin University, Geelong, VIC 3220, Australia (email: jihong.park@deakin.edu.au).}

  \footnotetext[3]{S. Oh and S.-L. Kim are with the School of Electrical \& Electronic Engineering, Yonsei University, 50 Yonsei-Ro, Seodaemun-Gu, Seoul 03722, Korea (email: seoh@ramo.yonsei.ac.kr, slkim@yonsei.ac.kr).}
  \footnotetext[4]{M. Bennis is with the Centre for Wireless Communications, University of Oulu, Oulu 90014, Finland (email: mehdi.bennis@oulu.fi).}
  \arabicfootnotes

Machine learning is one of the key building blocks in 5G and beyond \cite{park2018wireless,park2020extreme,park2020:cml} spanning a broad range of applications and use cases. In the context of mission-critical applications \cite{park2020extreme,MehdiURLLC:18}, machine learning models should be trained with fresh data samples that are generated by and dispersed across edge devices (e.g., phones, cars, access points, etc.). Collecting these raw data incurs significant communication overhead, which may violate data privacy. In this regard, \emph{federated learning (FL)} \cite{McMahan2016,Sumudu:20,Kim:CL20,Google:FL19} is a promising communication-efficient and privacy-preserving solution that periodically exchanges local model parameters, without sharing raw data. However, exchanging model parameters is extremely costly under modern deep neural network (NN) architectures that often have a huge number of model parameters. For instance, {MobileBERT is a state-of-the-art NN architecture for on-device natural language processing (NLP) tasks, with 25 million parameters corresponding to 96 MB \cite{sun2020mobilebert}. Training such a model by exchanging the 96 MB payload per communication round is challenging particularly under limited wireless resources. }

\begin{figure}
	\centering
	\includegraphics[width=\textwidth]{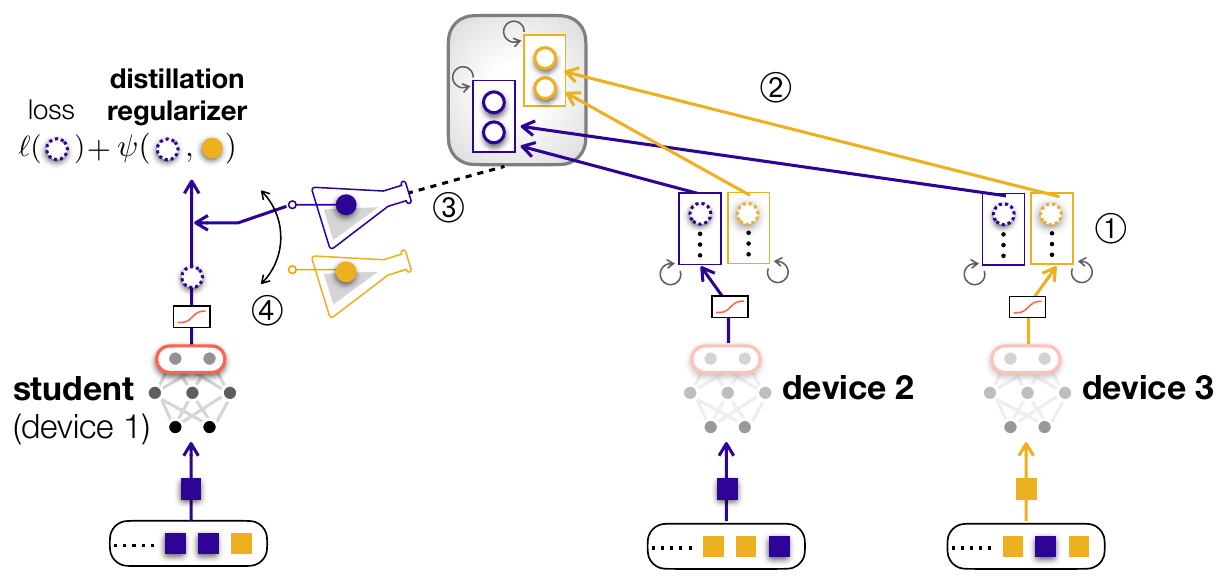}
	\caption{\small{A schematic illustration of federated distillation (FD) with 3 devices and 2 labels in a classification task.}} \label{Fig:Overview_FD}
	\end{figure}

The aforementioned limitation of FL has motivated to the development of \emph{federated distillation (FD)} \cite{MLPCD} based on exchanging only the local model outputs whose dimensions are commonly much smaller than the model sizes (e.g., 10 labels in the MNIST dataset). To illustrate, as shown in Figure~\ref{Fig:Overview_FD}, consider a 2-label classification example wherein each worker in FD runs local iterations with samples having either blue or yellow ground-truth label. For each training sample, the worker generates its prediction output distribution, termed a \emph{local logit} that is a softmax output vector of the last NN layer activations (e.g., $\{\text{blue},\text{yellow}\}=\{0.7,0.3\}$ for a blue sample). At a regular interval, the generated local logits of the worker are averaged per ground-truth label, and uploaded to a parameter server for aggregating and globally averaging the \emph{local average logits} across workers per ground-truth label. The resultant \emph{global average logits} per ground-truth label are downloaded by each worker. Finally, to transfer the downloaded global knowledge into local models, each worker updates its model parameters by minimizing its own loss function, in addition to a regularizer that penalizes larger gap between its own logit of a given sample and the global average logit for the given sample's ground-truth.

The overarching goal of this chapter is to provide a deep understanding of FD and show the effectiveness of FD as a communication-efficient distributed learning framework that is applicable to a variety of tasks. To this end, the rest of this chapter is organized into three parts. To demystify the operational principle of FD, by exploiting the theory of neural tangent kernel (NTK) \cite{NTK}, the first part in Chapter~\ref{Chapt:Preliminaries} provides a novel asymptotic analysis for two foundational algorithms of FD, namely knowledge distillation (KD) and co-distillation (CD). Next, the second part in Chapter~\ref{Chapt:FD} elaborates on a baseline implementation of FD for a classification task, and illustrates its performance in terms of accuracy and communication efficiency compared to FL. Lastly, to demonstrate the applicability of FD to various distributed learning tasks and environments, the third part presents two selected applications, namely FD over asymmetric uplink-and-downlink wireless channels and FD for reinforcement learning in Chapters~\ref{Chapt:FLD} and ~\ref{Chapt:FRD}, respectively, followed by concluding remarks in Chapter~\ref{Chapt:Conclusion}

  \section{Preliminaries: Knowledge Distillation and Co-Distillation} \label{Chapt:Preliminaries}

FD is built upon two basic algorithms. One is KD that transfers a pre-trained teacher model's knowledge into a student model \cite{HintonKD}, whereas the other is an online version of KD without pre-training the teacher model, called CD \cite{HintonCD}. Although KD has widely been used in practice since its inception, its fundamentals have not been fully understood up until now. Only a handful works \cite{AnalysisKD,phuong19a,tang2020understanding} have attempted to analyze KD and its convergence, using the recently proposed NTK technique \cite{NTK} as we will review in the first part of this section. Leveraging and extending this NTK framework, in the second part, we will provide a novel NTK analysis of the convergence of CD.

\begin{figure}\label{Fig:KD}
	\centering
	\includegraphics[width=.7\linewidth]{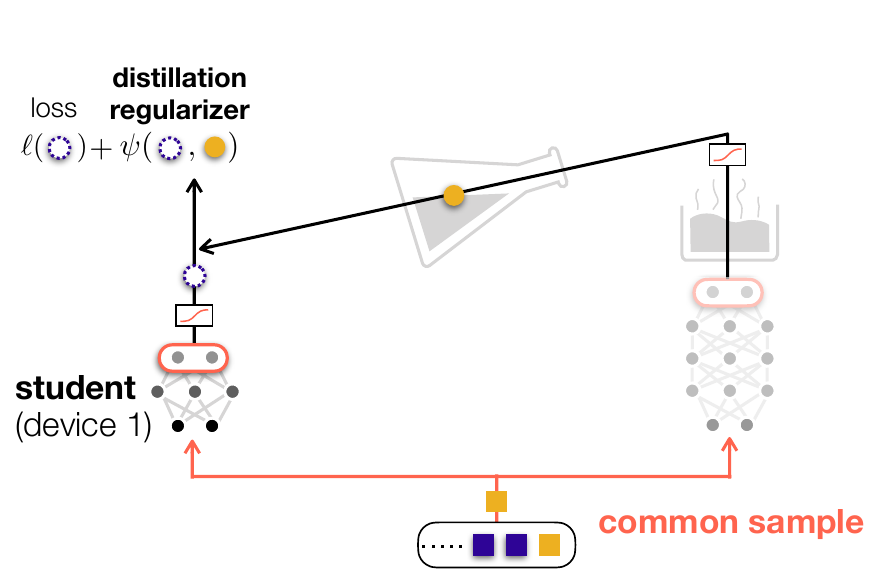}
	\caption{A schematic illustration of knowledge distillation (KD) from a pre-trained teacher to a student model.}
\end{figure}

\subsection{Knowledge Distillation}
Knowledge distillation (KD) aims to imbue an empty student model with a teacher’s knowledge \cite{HintonKD}. In a classification task, KD is different from the standard model training that attempts to match a target model's one-hot prediction (e.g., [cat, dog] = [0,1]) of each unlabeled sample with its ground-truth label. Instead, KD tries to match the target model’s output layer activation, i.e., logit\footnotemark (e.g., [cat, dog] = [0.3, 0.7]), with the teacher’s logit for the same sample. This logit contains more information than its one-hot prediction, thereby training the student model faster than the standard training with much less samples~\cite{phuong19a}. 

\footnotetext{
	{KD originally aims to match the softmax activation function of the student's logit with	the temperature softmax activation function of the teacher's logit \cite{HintonKD}. Recent KD works have also considered various activation functions of logits, such as margin rectifier linear unit (ReLU) and attention \cite{heo2019overhaul}. In this chapter, we consider the same activation functions as in~\cite{HintonKD}, and for the sake of convenience we hereafter call this functional output as logit.}
}

The teacher’s knowledge of KD can be constructed in various ways. Typically, the knowledge is a pre-trained teacher model’s logit, which is transferred to a small-sized student model for model compression \cite{HintonKD}. The knowledge can also be an ensemble of other student models’ logits \cite{HintonCD}, in that the ensemble of predictions is often more accurate than individual predictions. Leveraging this, one can train a student model by transferring the ensemble of other student models' logits. Indeed, CD and FD utilize this key idea for enabling KD-based distributed learning without the need for any pre-training operations, to be elaborated in Chapters~\ref{Chapt:CD} and \ref{Chapt:FD}.

Given the aforementioned teacher's knowledge, what the student model knows after KD can be clarified through the lens of NTK, a recently developed kernel method to asymptotically analyze an over-parameterized NN in an infinite width regime \cite{NTK}. To illustrate, we consider a simple 3-layer student NN model comprising input, hidden, and output layers with $M_i>0$, $M_h\rightarrow\infty$ (i.e., infinite width), and $M_o=1$ neurons, respectively. These layers are fully connected, and a non-linear activation function is applied to the hidden layer. In a classification task, the input data tuple $\{(\mathbf{x}_i, y_i)\}_{i = 1}^{n}$ consists of an unlabeled data sample $\mathbf{x}_i$ and its ground-truth label $y_i$. For a given input sample $\mathbf{x}_i$, the prediction output~$\hat{y}_i$ of the student NN is represented by the function $f(\mathbf{x}_i)$ as follows:
\begin{align}\label{eq:estimate}
\hat{y}_i = f(\mathbf{x}_i) = {\frac{1}{\sqrt{M_h}}} \sum_{m = 1}^{M_h} a_m f_m(\mathbf{x}_i) \in \mathbb{R},
\end{align}
where $\sigma(\cdot)$ is a real and non-linear activation function, $f_m(\mathbf{x}_i)=\sigma(\mathbf{w}_m^T \mathbf{x}_i)$ is the $m$-th activation of the hidden layer, and $\{\mathbf{w}_m\}_{m = 1}^{M_h}$ are the weights connecting the input and hidden layers. For the given NN architecture, the logit vector is the hidden layer activations $\{f_m(\mathbf{x}_i)\}_{m=1}^{M_h}$, of which the entries are linearly combined with the weight parameters $\{a_m\}$, resulting in the prediction output $\hat{y}_i$ of the student model.

In KD, the student model updates its weights $\{\mathbf{w}_m\}$ by minimizing its own loss function and a distillation regularizer that penalizes the student when the logit gap between the student and teacher is large. Applying the mean squared error function\footnotemark to both loss function and regularizer, the problem of KD is cast~as:
\begin{align} \label{Eq:KD_problem}
	\min_{\{\mathbf{w}_m\}} \underbrace{\sum_i (y_i - \hat{y}_i)^2}_{\text{loss}} + \lambda \underbrace{\sum_i \sum_m (\phi_m(\mathbf{x}_i) - f_m(\mathbf{x}_i))^2}_{\text{distillation regularizer}},
	\end{align}
	where $\lambda > 0$ is a constant hyperparameter and $\{\phi_m(\mathbf{x}_i)\}_{m=1}^{M_h}$ are pre-trained teacher model's logits and $\{f_m(\mathbf{x}_i)\}_{m=1}^{M_h}$ are student's logits. Note the number of logits at both teacher and student are assumed to be the same.
	
	\footnotetext{ { In KD under classification tasks, it is common to use the cross entropy functions for the loss and distillation regularizer. For the sake of the mathematical tractability, following \cite{AnalysisKD}, we consider the mean squared error functions for the loss and regularizer during the NTK analysis, while considering the cross entropy functions for the rest of this chapter.}}

To solve the problem \eqref{Eq:KD_problem}, following the standard NTK settings \cite{NTK,AnalysisKD}, we use the gradient descent algorithm with an infinitesimal step size. This results in the convergence of a trajectory of the discrete algorithm to a smooth curve modeled by a continuous-time differential equation as
\begin{align}\label{eq:dynamics_weight}
\frac{d}{dt}\mathbf{w}_m(t) =  \mathbf{L}_m(t)\left[\frac{a_m}{\sqrt{M_h}}(\mathbf{y} - \hat{\mathbf{y}}(t)) + \lambda (\boldsymbol{\phi}_m - \mathbf{f}_m(t))\right],
\end{align}
where $\mathbf{y}$ and $\hat{\mathbf{y}}(t)$ are respectively the vectors of the ground truth labels and the prediction outputs at time $t$, and $\boldsymbol{\phi}_m$ and $\mathbf{f}_m(t)$ are respectively the vectors of the teacher model's $m$-th logit and the student model's $m$-th logit at time $t$. The matrix $\mathbf{L}_m(t)$ consists of $\sigma'(\mathbf{w}_m^\mathsf{T} (t)\mathbf{x}_i)\mathbf{x}_i$ as its $i$-th column, where $\sigma'(\cdot)$ is the first derivative of the activation, which is also assumed to be Lipschitz continuous.

Generally, the dynamics of the weights described in \eqref{eq:dynamics_weight} are hard to analyze, yet we can still analyze the dynamics of the logits based on the following relation:
\begin{align}
\frac{d}{dt}\mathbf{f}_m(t) &= \mathbf{L}_m^\mathsf{T}(t) \frac{d}{dt}\mathbf{w}_m(t)\\
&= \mathbf{H}_m(t)\left[\frac{a_m}{\sqrt{M_h}}(\mathbf{y} - \hat{\mathbf{y}}(t)) + \lambda (\boldsymbol{\phi}_m - \mathbf{f}_m(t))\right],\label{eq:KDdynamics1}
\end{align}
{where $\mathbf{H}_m(t) = \mathbf{L}_m^\mathsf{T}(t) \mathbf{L}_m(t)$ is often called an NTK \cite{NTK}.}

{Empirically, in a network with a large number of parameters, it is observed that every weight vector along the trajectory of gradient descent algorithm is static over time and stays very close to its initialization. Based on such interesting observation, the theory of NTK establishes that the over-parametrization and random initialization jointly induce a \emph{kernel regime}, i.e., $\mathbf{H}_m(t) \approx \mathbf{H}_m(0)$ for $t \geq 0$ \cite{NTK,GDprove}, thereby giving rise to simpler dynamics under the negligible effect of $\mathbf{H}_m(t)$ on \eqref{eq:KDdynamics1}.
\begin{remark}[Theorem 1 in \cite{AnalysisKD}]
In the kernel regime, under mild assumptions on the eigenvalues of the matrices $\{\mathbf{H}_m(0)\}_{m =1}^{M_h}$ at initialization, bounded inputs and bounded weights, it can be shown that the student NN output vector $\mathbf{f}(t)$, which is the vector of $\{f(\mathbf{x}_i)\}_{i=1}^n$, converges asymptotically as
\begin{align} \label{Eq:NTK_KD}
\lim_{t \rightarrow \infty} \mathbf{f}(t) = \mathbf{f}_{\infty} = \frac{1}{a + \lambda}\left(a\mathbf{y} + \lambda \sum_{m = 1}^{M_h} \frac{a_m \phi_m}{\sqrt{M_h}} \right).
\end{align}
\end{remark}
\begin{proof}
Based on the observation that the behavior of gradient descent on the over-parametrized NN can be approximated by a linear dynamics of finite order, the evolution of $\mathbf{f}(t)$ can be expressed as
\begin{align}
\mathbf{f}(t) = \mathbf{f}_{\infty} + \mathbf{u}_1e^{-p_d t} +\mathbf{u}_1e^{-p_d t}+\cdots+\mathbf{u}_d e^{-p_d t},
\end{align}
where $d$ is the order of the linear system, and complex-valued vectors $\mathbf{u}_1,\dots\mathbf{u}_d$ are determined by the dynamics. Moreover, the non-zero complex-values $p_1,\dots,p_d$ are the poles that correspond to the singular points of the Laplace transform of $\mathbf{f}(t)$. In \cite{AnalysisKD}, it is shown in detail that all existing poles are positive-valued under mild assumptions, such that $\mathbf{f}(t) \rightarrow \mathbf{f}_{\infty}$ for $t \rightarrow \infty$.
\end{proof}
}
Consequently, as shown by \eqref{Eq:NTK_KD}, the student model after KD outputs a weighted sum of the ground truth $\mathbf{y}$ and the teacher's prediction $\sum_{m} a_m\phi_m$. Then, the student's prediction error compared to $\mathbf{y}$ can be represented as
\begin{align} \label{Eq:KD_error}
	|| \mathbf{f}_\infty - \mathbf{y} ||_2 = \frac{\lambda}{a + \lambda}\left\lVert \mathbf{y} - \sum_{m} \frac{a_m \phi_m}{\sqrt{M_h}}  \right\rVert_2 .
\end{align}
This implies that the student's prediction error decreases as the pre-trained teacher's prediction $\sum_{m} \frac{a_m \phi_m}{\sqrt{M_h}} $ approaches to $\mathbf{y}$, i.e., an ideally trained teacher.

\begin{figure}\label{Fig:CD}
	\centering
	\includegraphics[width=.9\linewidth]{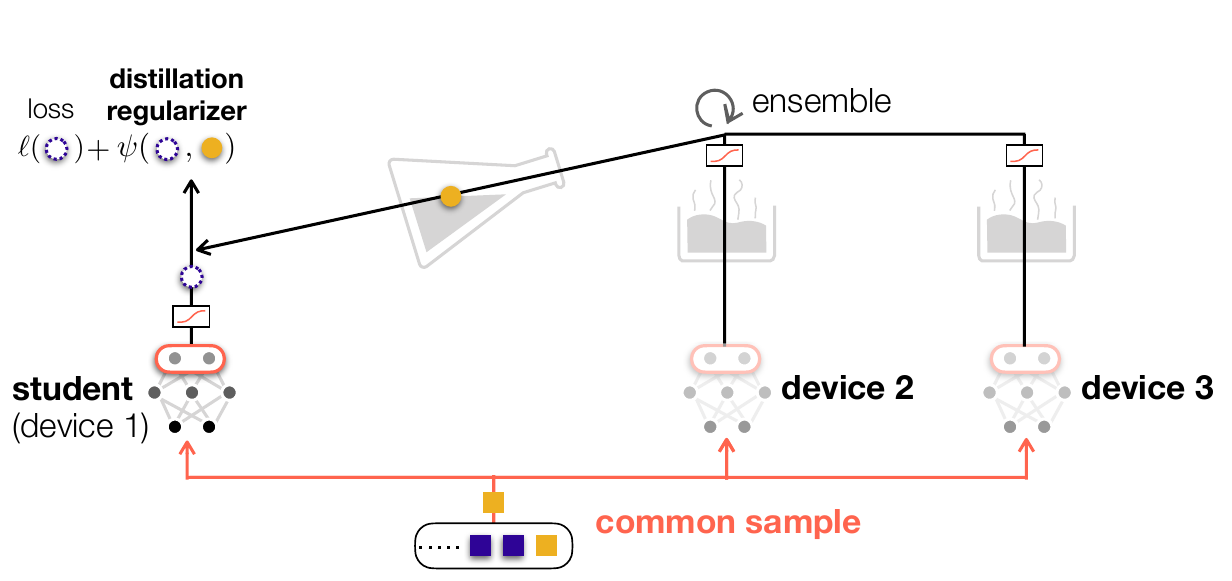}
	\caption{A schematic illustration of co-distillation (CD) among 3 student models without any pre-trained teacher model.}
\end{figure}

\subsection{Co-Distillation} \label{Chapt:CD}

KD postulates a pre-trained teacher model that hinders distributed learning operations. However, CD, which is an online version of KD, obviates the need for the pre-trained teacher model \cite{HintonCD}. The key idea of CD is to treat an ensemble of multiple models' prediction outputs as the teacher's knowledge, which is often more accurate than the individual prediction outputs \cite{GoodfellowBook:16,HintonCD}. To this end, each worker, i.e., student model, sees the ensemble of the other $C-1$ workers as a virtual teacher. Consequently, the problem of CD is given by recasting the problem \eqref{Eq:KD_problem} of KD as follows:
\begin{align} \label{Eq:CD_problem}
	\min_{\{\mathbf{w}^1_m\},\dots,\{\mathbf{w}_m^C\}} \sum_c  \Bigg( \underbrace{\sum_i(y_i - \hat{y}^c_i)^2}_{\text{loss}} + \lambda \underbrace{\sum_i \sum_m \Bigg(\frac{1}{C-1}\sum_{c' \neq c} f^{c'}_m(\mathbf{x}_i) - f^c_m(\mathbf{x}_i)\Bigg)^2}_{\text{distillation regularizer}}  \Bigg),
	\end{align}
	where $\hat{y}_i^c$ is the prediction output of the $c$-th worker, $\{\mathbf{w}_m^c\}$ is its weight parameters, and $\{f^c_m(\cdot)\}$ is its logits. Here, the pre-trained teacher's logit $\phi_m(\mathbf{x}_i)$ of KD in \eqref{Eq:KD_problem} is replaced with the ensemble logit $\frac{1}{C-1}\sum_{c' \neq c} f^{c'}_m(\mathbf{x}_i)$ of $C-1$ workers in CD. Note that the problem \eqref{Eq:CD_problem} of CD is formulated for all $C$ workers, rather than considering each worker separately. This problem is more challenging than KD, in that the teacher's knowledge becomes dependent on each worker (due to exclusion) and all the other workers (due to averaging). 

{
\begin{remark}\label{rem:CD}
Given the aforementioned interactions across workers, based on analysis in the kernel regime, it can be shown that the output of the workers converges to the ground-truth asymptotically as
\begin{align} \label{Eq:NTK_CD}
\lim_{r \rightarrow \infty} \mathbf{f}^c(r) = \mathbf{y},
\end{align}
for all $c \in \{1,\dots,C\}$, where $\mathbf{f}^c(r)$ is the output of the worker $c$ after local training with $r$-th global update (or communication round).
\end{remark}
\begin{proof}
Without loss of generality, we hereafter focus only on the first worker out of $C$ workers whose models are identically structured and independently initialized. After initialization and local training for warm-up, the workers share the first updates, i.e., $\mathbf{f}^1(0), \dots, \mathbf{f}^C(0)$. Then, each worker locally and iteratively runs GD with regularization until convergence. According to the result \eqref{Eq:NTK_KD} from KD, the output of the worker $1$ converges to 
\begin{align}
\mathbf{f}^1(1) = \frac{1}{a + \lambda}\left(a\mathbf{y}+\frac{\lambda}{C-1}\sum_{c = 2}^C \mathbf{f}^c(0) \right).
\end{align}
Thus, the output of the model $\mathcal{C}_1$ after $r$-th updates will converge to
\begin{align}
\mathbf{f}^1(r) &= \frac{1}{a + \lambda}\left(a\mathbf{y}+\frac{\lambda}{C-1}\sum_{c = 2}^C \mathbf{f}^c(r-1) \right)\\
&= \frac{1}{a + \lambda}\left(a\mathbf{y}+\lambda \frac{\sum_{c = 2}^C \left(a\mathbf{y} + \frac{\lambda}{C-1} \sum_{c' \neq c} \mathbf{f}^{c'}(r-2)\right)}{(C-1)(a+\lambda)} \right)\\
&= \frac{1}{a \! + \! \lambda}\left(a\mathbf{y}+\lambda\frac{(C\!-\!1)a\mathbf{y} + \lambda\mathbf{f}^1(r\!-\!2) + \frac{\lambda(C-2)}{C-1} \sum_{c = 2}^{C} \mathbf{f}^c(r\!-\!2)}{(C-1)(a+\lambda)} \right).\label{eq:ithupdate}
\end{align}
By introducing $\mathbf{v}_{r+1} = \frac{\lambda}{C-1}\sum_{c = 2}^{C} \mathbf{f}^c(r) = (a+\lambda)\mathbf{f}^1(r+1) - a\mathbf{y}$, we can simplify \eqref{eq:ithupdate} to a linear non-homogeneous recurrence relation:
\begin{align}
\mathbf{v}_{r+1} \! = \! \frac{(C-2)\lambda}{(C\!-\!1)(a\!+\!\lambda)} \mathbf{v}_{r} \! +\! \frac{\lambda^2}{(C\!-\!1)(a\!+\!\lambda)^2} \mathbf{v}_{r-1} \!+\! \frac{\lambda^2 a\mathbf{y}}{(C\!-\!1)(a\!+\!\lambda)^2}\! +\! \frac{\lambda a \mathbf{y}}{(a\!+\!\lambda)},
\end{align}
for $r \geq 1$. By solving the above recurrence relation \cite{Rosen:book}, we obtain the closed-form solution 
\begin{align}	
\mathbf{v}_r = \alpha \left(\frac{\lambda}{a+\lambda}\right)^r + \beta \left(-\frac{\lambda}{(C-1)(a+\lambda)}\right)^r + \lambda \mathbf{y},\label{EQ:NTK_CD_interim}
\end{align}
where $\alpha = \frac{\lambda}{C}\sum_{c = 1}^{C} \mathbf{f}^c(0) - \lambda \mathbf{y}$ and $\beta = \frac{\lambda}{C(C-1)}\sum_{c = 2}^{C} \mathbf{f}^c(0) - \frac{\lambda}{C} \mathbf{f}^{1}(0)$. Note that for $r \rightarrow \infty$,
\begin{align}
\lim_{r \rightarrow \infty} \mathbf{v}_r = \lambda\mathbf{y},
\end{align}
since $|\frac{\lambda}{a+\lambda}|< 1$ and $|-\frac{\lambda}{(C-1)(a+\lambda)}|<1$ for $C \geq 2$. Consequently, we can see that the output of the worker $1$ converges to the ground-truth as
\begin{align} \label{Eq:NTK_CD}
\lim_{r \rightarrow \infty} \mathbf{f}^1(r) &= \frac{1}{a+\lambda}\left(a\mathbf{y} + \lambda \mathbf{y}\right) = \mathbf{y}.
\end{align}
In the same way, the result \eqref{Eq:NTK_CD} of the worker $1$ can be extended to any worker with the same conclusion. This ends the proof of Remark \ref{rem:CD}.
\end{proof}
}
Such a result in Remark \ref{rem:CD} is remarkable in that CD achieves zero prediction error that is achievable under KD only when the teacher model is ideally pre-trained as shown in \eqref{Eq:KD_error}. This result highlights the importance of continual training that allows workers to reach the maximum prediction capability, as opposed to KD that is additionally guided by a pre-trained yet fixed teacher model.

\begin{figure}[t] 
	\centering
	\includegraphics[width=.7\linewidth]{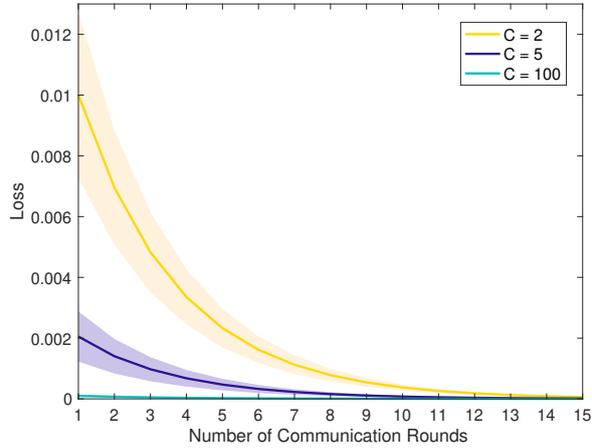} 
	\caption{Learning curves of CD with $C \in\{2, 5, 100\}$ workers, capturing the loss converging to 0 as the number $r$ of communication rounds increases.} \label{Fig:CD_sim}
\end{figure}

Lastly, it is notable that more workers yield faster convergence of CD. In essence, the convergence is achieved by eliminating the first two terms in the RHS of \eqref{EQ:NTK_CD_interim}. These two terms decrease not only with the number of communication rounds $r$ but also with the number of workers $C$. This implies that with more workers one needs less communications until convergence. Furthermore, we conceive that when $C\rightarrow\infty$, only one communication round can achieve convergence, enabling one-shot CD. 

Fig.~\ref{Fig:CD_sim} corroborates the aforementioned theoretical results by numerical evaluations of CD for a simple classification task considering $10$ classes of samples labeled $0$-$9$, generated with an arbitrary mapping function which is unknown to the workers. The result shows that as the number $C$ of workers grows, the convergence speed of CD increases while the variance reduces. Furthermore, as expected by the theoretical result in \eqref{Eq:NTK_CD}, numerical simulations validate that even with $C=2$, CD is guaranteed to converge. Lastly, for $C=100$, one can achieve convergence with only one communication round $r=1$, verifying the feasibility of one-shot~CD.

  \section{Federated Distillation} \label{Chapt:FD}

CD has a great potential in enabling fast distributed learning with high accuracy as demonstrated in the previous section, yet its communication efficiency is still questionable. The fundamental reason traces back to KD that requires common training sample observations by both student and teacher models. For an online version of KD, this implies that all workers should observe the same sample per each loss calculation, requiring extensive sample exchanges that may also violate local data privacy. Eliminating such a dependency on common sample observations is the key motivation for developing FD, as elaborated next.

\subsection{Federated Distillation for Classification}

In a classification task, FD avoids the aforementioned problem of common sample observations in CD by grouping samples according to labels, thereby extending CD to a communication-efficient distributed learning framework. As depicted by Figure~\ref{Fig:Overview_FD}, the operations of FD are summarized by the following four steps.
\begin{enumerate}
	\item Each worker stores a mean logit vector per label during local training.
	\item Each worker periodically uploads its \emph{local-average logit vectors} to a parameter server averaging the uploaded local-average logit vectors from all workers separately for each label. 
	\item Each worker downloads the constructed \emph{global-average logit vectors} of all labels from the server.
	\item During local training based on KD, each worker selects its teacher's logit as the downloaded global-average logit associated with the same label as the current training sample's ground-truth label.
\end{enumerate}

In what follows we describe the details of FD operations.
Similar to CD, we consider that the worker $c \in \{1,\dots,C\}$ has $n$ observed samples with ground-truth label, i.e., $\{(\mathbf{x}^c_i, y^c_i)\}_{i = 1}^{n}$, but independently observed at each worker. For the sake of simplicity, assume $\mathcal{Y}=\{1, 2, \dots, |\mathcal{Y}|\}$ to be an alphabet of $|\mathcal{Y}|$ labels under consideration, and define an index set $\mathcal{I}^c_{\ell}$, that is composed of $\ell$-labeled sample indices at the worker $c$, where $|\mathcal{I}_\ell^{c}| = n_\ell^c$ and $\sum_{\ell}n_\ell^c = n$. Under such circumstances, FD aims to solve the following optimization~problem:
\begin{align} \label{Eq:FD}
	\min_{\{\mathbf{w}^1_m\},\dots,\{\mathbf{w}_m^C\}} \sum_c  \sum_{\ell} \Bigg( \underbrace{\sum_{i \in \mathcal{I}_\ell^c }(y_i - \hat{y}^c_i)^2}_{\text{loss}} + \lambda \underbrace{\sum_{i \in \mathcal{I}_{\ell}^{c}} \sum_m \Bigg(\frac{1}{C-1}\sum_{c' \neq c} \bar{f}_{m,l}^{c'} - f^c_m(\mathbf{x}^c_{i})\Bigg)^2}_{\text{distillation regularizer}}  \Bigg),
\end{align}
where $f^c_m(\cdot)$ is the $m$-the logit of the worker $c$ as before, $\bar{f}^c_{m,l} = \frac{1}{n_l^{c}}\sum_{i \in \mathcal{I}_{\ell}^{c}} f^{c}_m(\mathbf{x}_{i}^c)$ is the local average of the worker $c$'s $m$-th logit for the samples labeled $l$.

\begin{algorithm}[t]{
	\small
		\caption{Federated Distillation (FD)}\label{euclid}
		\begin{algorithmic}[1]
		 \Require Prediction: $f(\mathbf{x})$, Ground-truth label: $y$, Loss function: $\mathcal{L}(f(\mathbf{x}), y)$
		
				  \While  {not converged}
						\Procedure{Local Training Phase }{at worker $\forall c \in \{1,\dots,C\}$}
				\For  {$k$ steps}
				: $\mathcal{B}$ $\gets$ $\mathcal{S}^c$
				\For  {sample $\mathbf{x}_b$ and label $y_b$, for $b\in \mathcal{B}$}
				 \State$\mathbf{w}^{c} \gets \mathbf{w}^{c}-\eta \nabla \{ \mathcal{L}(f^c(\mathbf{x}_b), y_b) + \lambda \cdot \mathcal{L}(F^c(\mathbf{x}_b), \hat{F}_{y_b,r}^{c}) \}$
				\State  $F^{c}_{y_{b},r} \gets F^{c}_{y_{b},r}+F^c(\mathbf{x}_b)$, $\mathsf{cnt}^{c}_{y_{b},r} \gets \mathsf{cnt}^{c}_{y_{b},r}+1$
		
				\EndFor 
				\EndFor

				\For {label $\ell =1,2,\cdots,|\mathcal{Y}|$}	        
				\State $\bar{F}_{\ell,r}^{c} \gets F_{\ell,r}^{c} / \mathsf{cnt}^{c}_{\ell,r}:$ 
				\Return {$\bar{F}_{\ell,r}^{c}$} to server
				\EndFor   
				\EndProcedure
			\Procedure  {Global Ensembling Phase } {at the server}
			\For  {each worker $c=1,2,\cdots,C$}
			\For  {label $\ell=1,2,\cdots,|\mathcal{Y}|$}
			\State  $\bar{F}_{\ell,r} \gets \bar{F}_{\ell,r} + \bar{F}_{\ell,r}^{c}$
			\EndFor
			\EndFor
			\For  {each worker $c =1,2,\cdots,C$}
			\For  {label $\ell=1,2,\cdots,|\mathcal{Y}|$}
			\State  $\hat{F}_{\ell,r+1}^{c} \gets \bar{F}_{\ell,r} - \bar{F}_{\ell,r}^{c}$, $\hat{F}_{\ell,r+1}^{c} \gets \frac{\hat{F}_{\ell,r+1}^{c}}{(C-1)}:$ \Return {$\hat{F}_{\ell,r}^{c}$} to worker $c$  
			\EndFor
			\EndFor
			\EndProcedure
	
			\EndWhile{end while}
			\end{algorithmic}}
		\end{algorithm}

Following the aforementioned four-step operations, FD solves the problem \eqref{Eq:FD} using Algorithm~\ref{euclid}. Notations are summarized as follows. The set $\mathcal{S}^c$ denotes the training dataset of the worker $c$, and $\mathcal{B}$ represents a set of sample indices drawn as a batch per worker during the local training phase. The function $F^c(\cdot)$ is a logit vector, made by vectorizing the logits $\{f^c_m(\cdot)\}$. The function $\mathcal{L}(p,q)$ is a quadratic loss function, measuring the mean squared error between $p$ and $q$, which is used for both loss function and distillation regularizer. Note that the quadratic loss can be replaced with any other well-defined loss function, such as cross-entropy. As opposed to the asymptotic analysis, we consider a constant learning rate $\eta$ for practicality, and $\lambda$ is a weighting constant for the distillation regularizer. At the $c$-th worker, $\bar{F}_{\ell,r}^{c}$ is the local-average logit vector at the $r$-th iteration when the training sample belongs to the $\ell$-th ground-truth label, $\hat{F}_{\ell,r}^{c}$ is the global-average logit vector that equals $\hat{F}_{\ell,r}^{c}=\sum_{c'\neq c}\bar{F}^{c'}_{\ell,r}/(C-1)$ with $C$ workers, and $\mathsf{cnt}^{c}_{\ell,c}$ counts the number of samples whose ground-truth label is $\ell$.

\begin{figure}[t] 
	\centering
	\includegraphics[width=.7\linewidth]{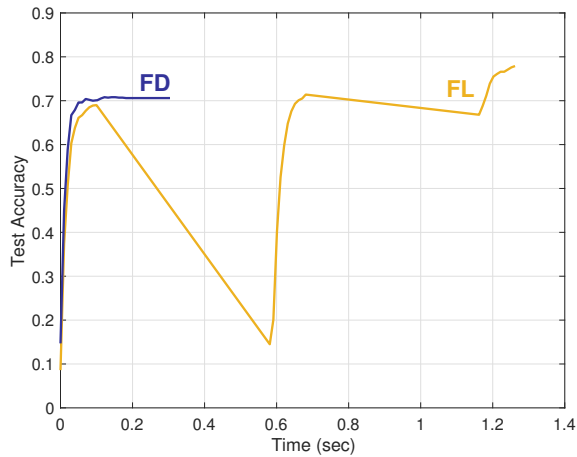}
	\caption{Learning curves of FD and FL with $2$ workers for the MNIST classification.} \label{Fig:FD_learning}
\end{figure}

\begin{figure}[t] 
	\centering
	\subfigure[Test accuracy.]{\includegraphics[width=.7\linewidth]{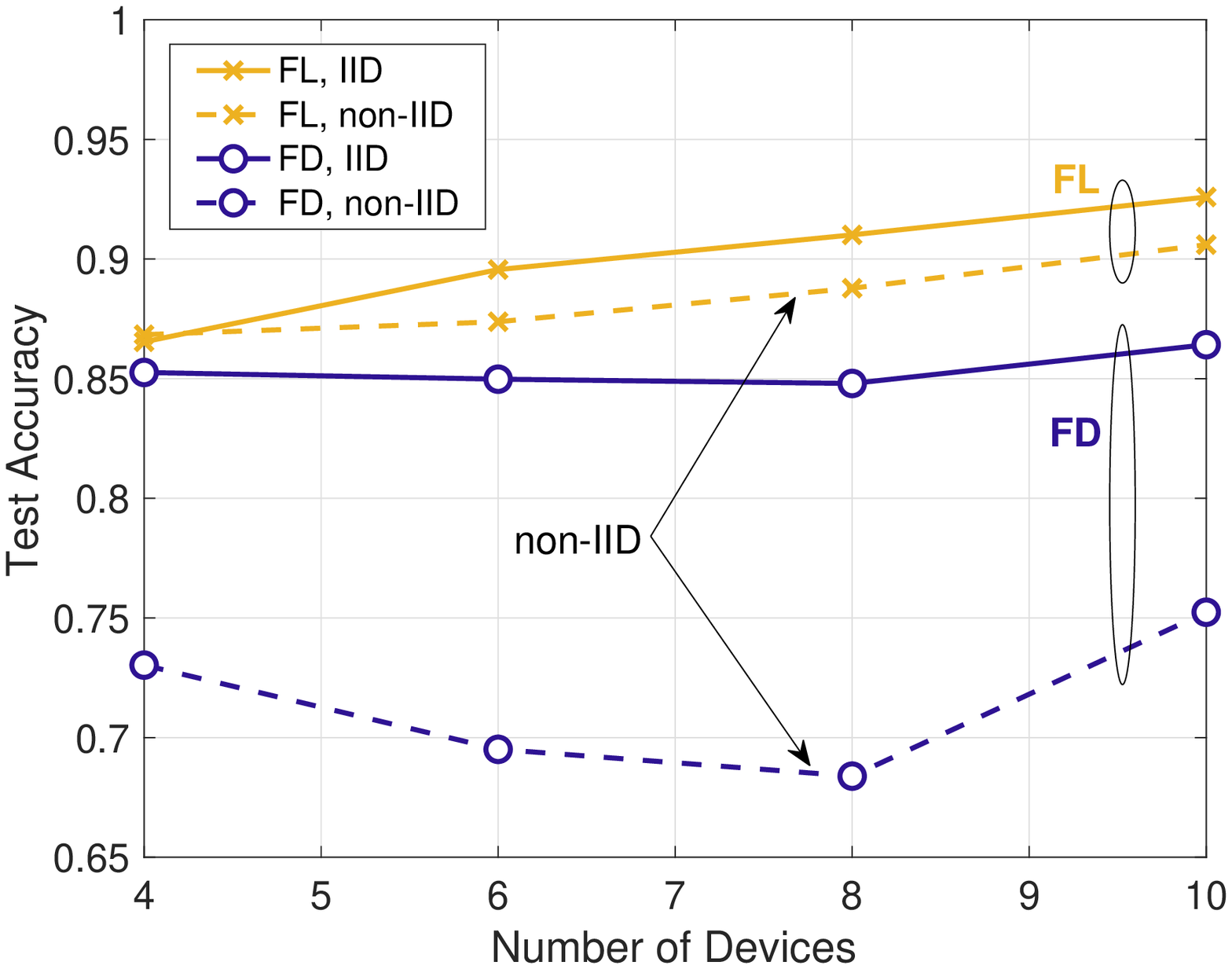}}
	\subfigure[Sum communication cost.]{\includegraphics[width=.7\linewidth]{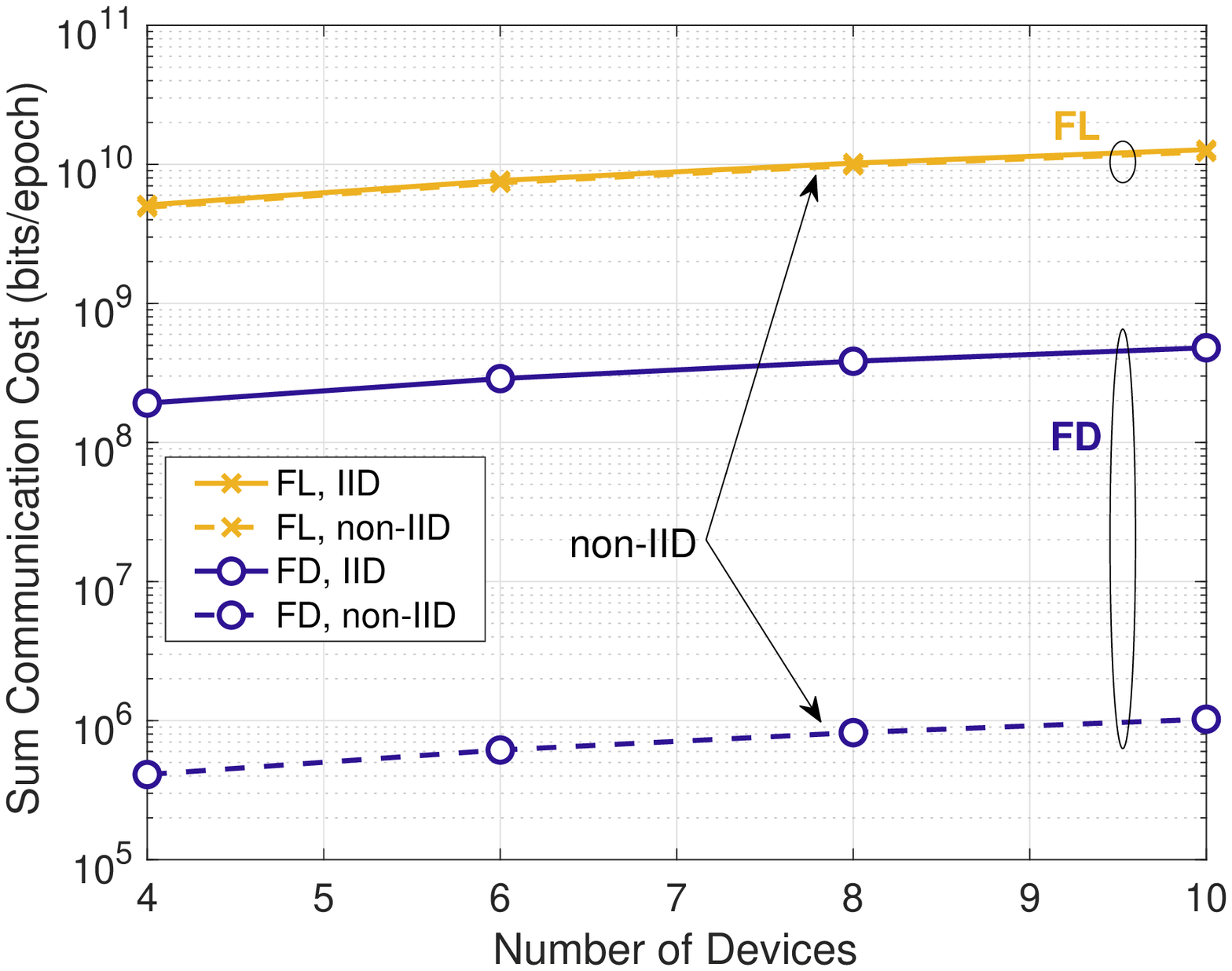}}
	\caption{Comparison between FD and FL in terms of (a) test accuracy and (b) sum communication cost of all workers per epoch, under an IID or non-IID MNIST data.} \label{Fig:FD_sim}
\end{figure}

Figure~\ref{Fig:FD_learning} shows the numerical evaluations of FD for the MNIST (hand-written $0$-$9$ images) classification task. The result illustrates that FD achieves $4.3$x faster convergence than FL while compromising less than $10$\% accuracy, under a $5$-layer convolutional NN operated by $2$ workers (see more details in \cite{MLPCD}. To see the effectiveness of FD in a more generic scenario, Figure~\ref{Fig:FD_sim} considers up to $10$ workers, and both cases of an independent and identically distributed (IID) local dataset and a non-IID dataset whose local data samples are imbalanced across labels. The result shows that for different numbers of workers, FD can always reduce around $10,\!000$x communication payload sizes per communication round compared to FL. Considering both fast convergence and payload size reduction, FD reduces the total communication cost until convergence by over $40,\!000$x compared to FL. Nonetheless, FD still comes at the cost of compromising accuracy, particularly under non-IID data distributions.

\subsection{Recent Progress and Future Direction}

The aforementioned implementation of Vanilla FD focuses only on reducing communication payload sizes in a classification task at the cost of sacrificing accuracy. Several recent works have substantiated the communication efficiency of FD under more realistic wireless environments without compromising accuracy for applications beyond classification as reviewed next.

\begin{itemize}
	\item  \emph{FD Over Wireless} --  
	FD is a communication-efficient distributed learning framework, as demonstrated by achieving $40,\!000$x less total communication cost than FL for an image classification task in the previous section. The communication efficiency of FD also holds under wireless fading channels \cite{MixFLD,Ahn,Takayuki}. Even with low signal-to-noise ratio and/or bandwidth, the payload size reduction of FD can be turned into more successful receptions and/or lower latency, resulting in even higher accuracy than FL \cite{Ahn,Takayuki}. It could be interesting to see the effectiveness of FD under more realistic wireless environments with advanced physical-layer and multiple-access techniques such as time-varying millimeter-wave channels, reconfigurable intelligent surfaces, non-orthogonal multiple access, and many more.

	\item \emph{Communication Efficiency vs. Accuracy} --
	FD is more vulnerable to the problem of non-IID data distributions compared to FL. Even if a worker obtains the global average logits for all labels, when the worker lacks samples of a specific target class, the global knowledge is rarely transferred into the worker's local model. Furthermore, in many cases \cite{MLPCD,MixFLD,Ahn,FRD}, the communication efficiency of FD comes at the cost of compromising accuracy, yielding the trade-off between FD and FL. Given the trade-off between FD's higher communication efficiency and FL's higher accuracy, it is possible to utilize both of their strengths by taking into account the nature of uplink-downlink asymmetric channels. As shown in \cite{FLD,MixFLD}, one can exploit FL in the downlink and FD in the uplink whose capacity is much less than the downlink due to the low transmission energy at the devices, to be further discussed in Chapt.~\ref{Chapt:FLD}. 

	\item \emph{Proxy Data Aided FD} --
	Recent works have overcome the aforementioned limitations of FD, i.e., accuracy degradation particularly under non-IID data distributions. The core idea is to additionally construct a common proxy dataset (e.g., a public dataset \cite{Ahn} or mean samples per label \cite{Takayuki}) through which the local KD operations and the local logits to be uploaded are provided. In fact, as opposed to FL that exchanges each worker's freshest model updated right before uploading, FD is based on exchanging the locally averaged logits during which each worker's model is progressively updated. To resolve this issue, workers can collectively construct a global proxy dataset by averaging all data samples per label, referred to as global average covariate vectors in \cite{Ahn} or by using a pre-arranged public dataset \cite{Takayuki}. Utilizing such a proxy dataset, one can generate the local logits to be exchanged right before uploading, thereby distilling the knowledge from the freshest models. Furthermore, operating KD through the proxy dataset makes all workers observe the same samples, thereby avoiding any possible errors induced by coarse sample grouping in the original FD. Consequently, as demonstrated in \cite{Takayuki}, such proxy dataset aided FD can achieve higher accuracy than FL even under non-IID local data distributions. Extending this line of research, it could be worth investigating how to construct the proxy dataset using a coreset, a small dataset approximating the original data distribution \cite{FLD,Coreset}.

	\item \emph{FD Beyond Classification} --
The applicability of FD is not limited to classification tasks in supervised learning. As shown by \cite{FRD}, FD can be applied to an reinforcement learning (RL) application by replacing the label-wise sample grouping of the original FD with clustering based on the neighboring states (e.g., locations) of RL agents, to be further elaborated in Chapt.~\ref{Chapt:FRD}. In unsupervised learning, it could be possible to collectively train multiple conditional generative adversarial networks (cGANs) \cite{CGAN} using FD by exchanging their discriminators' last layer activations that are grouped based on the common conditions of cGANs. Last but not least, in self-supervised learning, one could exploit FD to train multiple bootstrap your own latent (BYOL) networks, each of which comprises a pair of online and target models \cite{BYOL}, by constructing each target model's prediction based on an ensemble of the last layer activations of online models.

\end{itemize}

  \section{Application: FD Under Uplink-Downlink Asymmetric Channels} \label{Chapt:FLD}

Despite the communication efficiency brought by FD in the distributed learning framework, there still remains an accuracy issue especially under communication-limited scenarios. In a typical wireless communication network, the uplink communication is more limited by lower transmission power and smaller available bandwidth than the downlink~\cite{Park:16}, which we refer to as uplink-downlink channel asymmetry. Thus, for FD based distributed learning built over wireless networks, a large accuracy loss of model training is inevitable, since FD goes through a number of communication rounds for model training over both uplink and downlink channels.

In this context, as an advanced form of FD, the \emph{Mix2FLD} achieves both high accuracy and communication-efficiency under the uplink-downlink channel asymmetry. As depicted in Figure \ref{fig_Mix2FLD}, Mix2FLD is built upon two key algorithms: \emph{federated learning after distillation (FLD)}~\cite{FRD} and \emph{Mixup} data augmentation~\cite{Zhang2018}. Specifically, by leveraging FLD, each worker in Mix2FLD uploads its local model outputs as in FD, and downloads model parameters as in FL, thereby coping with the uplink-downlink channel asymmetry. Between the uplink and downlink, the server runs KD. However, this output-to-model conversion requires additional training samples collected from workers, which may violate local data privacy while incurring huge communication overhead.
To preserve data privacy with minimal communication overhead during seed sample collection, Mix2FLD utilizes a \emph{two-way Mixup algorithm (Mix2up)}, as illustrated in Figure \ref{fig_Mix2FLD}b. To hide raw samples, each worker in Mix2up uploads locally superposed samples using Mixup. Next, before running KD at the server, the uploaded mixed-up samples are superposed across different workers, in a way that the resulting sample labels are in the same form of raw sample labels. This inverse-Mixup provides more realistic synthetic seed samples for KD, without restoring raw samples. Furthermore, with the uploaded mixed-samples from the workers, a larger number of inversely mixed-up samples can be generated, thereby enabling KD with minimal uplink cost. In the following subsections, we first elaborate a baseline method, MixFLD that combines FLD and Mixup, followed by describing Mix2FLD that integrates MixFLD with the inverse-Mixup.




\subsection{Baseline: MixFLD}
MixFLD integrates FLD with Mixup, within which FLD counteracts the uplink-downlink channel asymmetry as elaborated next. Following FLD, as shown in Figure \ref{fig_Mix2FLD}a, at the $r$-th global update, the workers upload their local average logit vectors, thereby constructing a global average logit vector at the server, as in FD. Then, the workers download the global weight vector as in FL. To this end, the server must convert the global logit average vector into the global weight vector, since it lacks one. The key idea is to transfer the knowledge in the global average logit vector to a global model. To enable this, at the beginning of FLD, each worker uploads $n_{\text{mix}}$ seed samples randomly selected from its local dataset. By feeding the collected $Cn_{\text{mix}}$ seed samples, denoted by $\{\mathbf{x}_{s,i}\}_{i = 1}^{Cn_{\text{mix}}}$, the server runs $K_s$ iterations of SGD with KD, thereby updating the global model's weight vector $\mathbf{w}_{g,k}$~as:
\begin{align}
\mathbf{w}_{g,k+1} = \mathbf{w}_{g,k} - \eta \cdot \nabla \( \mathcal{L}(f_g(\mathbf{x}_{s,i}), \mathbf{y}_{s,i})  +  \lambda \cdot \mathcal{L}(F_{g}(\mathbf{x}_{s,i}) , \hat{F}_{y_{s,i},r}) \), \label{eq:FLD}
\end{align}
where $f_g(\cdot)$ is the function denoting the global NN at the server and $F_{g}$ is the corresponding global model logit vector. As defined earlier, $\mathcal{L}(\cdot,\cdot)$ is a well-designed loss function such as quadratic loss or cross-entropy and $\hat{F}_{l,r}$ is the global average logit vector for $l$-labeled samples at the $r$-th global update, which is obtained by averaging local logit vectors uploaded from the workers. Finally, the server yields the global model $\mathbf{w}_{g,K_s}$ that is downloaded by every worker. The remaining operations follow the same procedure of FL.

\begin{figure*}[t!]
	\centering
\subfigure[\small \textbf{Mix2FLD}: downlink federated learning (FL) \& uplink federated distillation~(FD) with two-way Mixup (Mix2up) seed sample collection.]{\includegraphics[width= .85\linewidth]{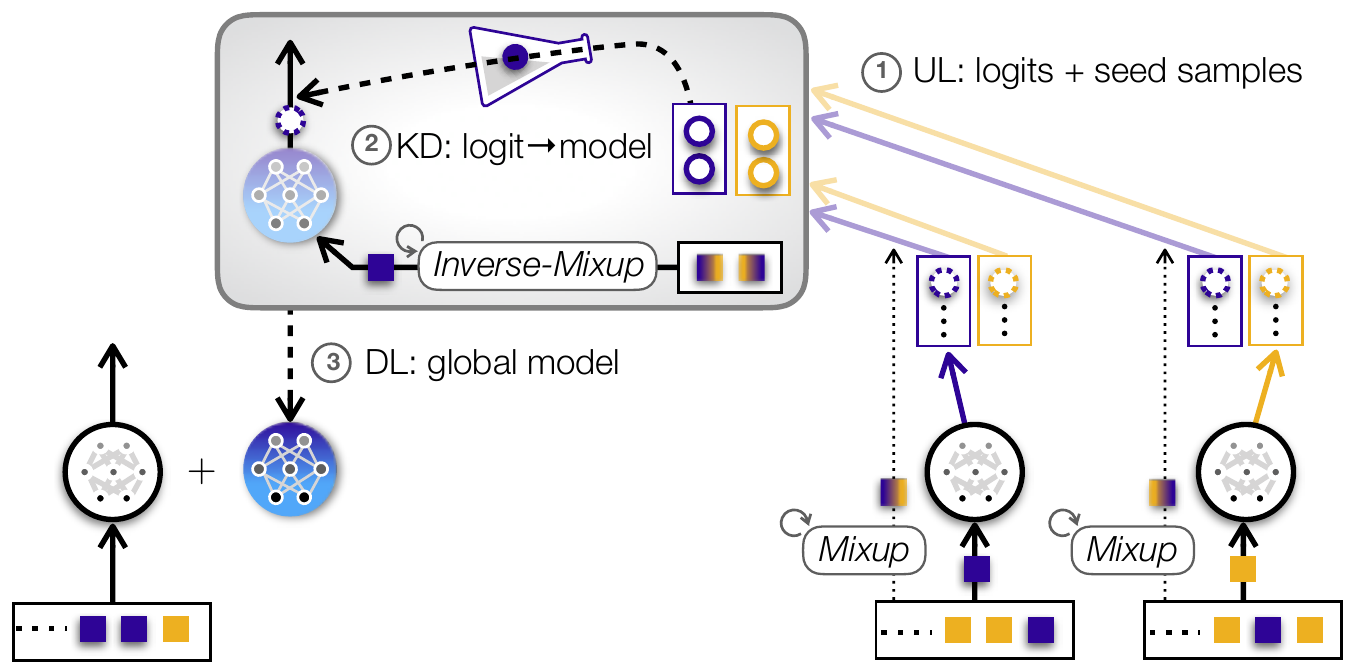}}
\subfigure[\small \textbf{Mix2up}: mixing raw samples at workers \& inversely mixing them across different workers at the server (mixing ratio {$\gamma=0.4$}).]{\includegraphics[width= .85\linewidth]{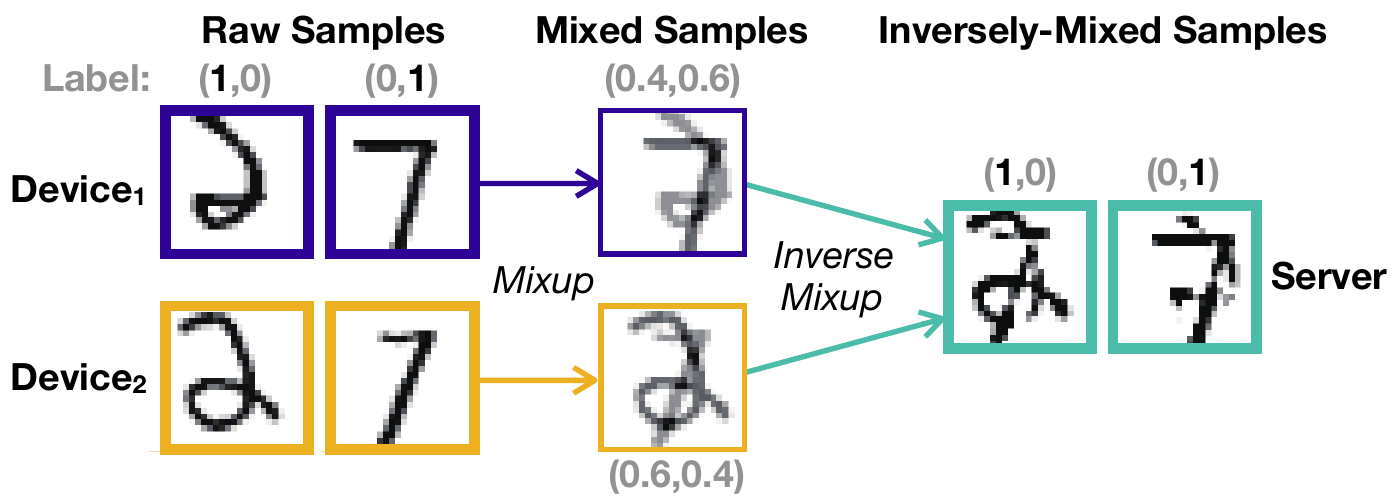}}
\caption{\small An illustration of (a) Mix2FLD operation and (b) Mix2up.}\label{fig_Mix2FLD}
\end{figure*}

The aforementioned FLD operations include seed sample collection process that may incur non-negligible communication overhead while violating local data privacy. To mitigate this problem, MixFLD applies Mixup before collection \cite{FLD,Zhang2018} to the sample collection procedure of FLD as follows. Before uploading the seed samples, the worker $c$ randomly selects two different raw samples $\mathbf{x}^c_i$ and $\mathbf{x}^c_j$ with $i\neq j$, having the ground-truth labels $y^c_i$ and $y^c_j$, respectively. With a mixing ratio $\gamma\in(0,0.5]$ given identically for all workers, the worker linearly combines these two samples (see Figure \label{fig_Mix2FLD}b), thereby generating a mixed-up sample $\hat{\mathbf{x}}^c_{ij}$ as:
\begin{align}
\hat{\mathbf{x}}^c_{ij} = \gamma \mathbf{x}^c_{i} + (1-\gamma) \mathbf{x}^c_{j}, \label{Eq:mixup}
\end{align}
whose label is also mixed up as $\hat{\mathbf{y}}_{ij}^c = \gamma \mathbf{y}^c_{i} + (1-\gamma) \mathbf{y}^c_{j}$. Then, each worker uploads the generated $n_{\text{mix}}$ mixed-up samples to the server without revealing raw samples.

{
	The guaranteed privacy level can be quantified through the lens of ($\varepsilon$,$\delta$)-differential privacy \cite{Dwork:DPBook}, in which lower $\epsilon,\delta>0$ preserves more privacy by making it difficult to guess with less confidence whether or not a certain data point is included in a private dataset. For the sake of the analysis, we consider that each worker selects two samples uniformly at random out of $n$ samples, and mixes them with $\gamma=0.5$, followed by inserting additive zero-mean Gaussian noises to $\hat{\mathbf{x}}^c_{ij}$ and $\hat{\mathbf{y}}^c_{ij}$ with the variances $\sigma_x^2$ and $\sigma_y^2$, respectively. When generating $n_{\text{mix}}$ samples at each worker, according to Theorem 3 in \cite{Lee:ISIT19}, the aforementioned Mixup is ($\varepsilon$,$\delta$)-differentially private where
\begin{align}
	\varepsilon = \frac{2 n_{\text{mix}} \Delta^2}{8n}\(1 + \sqrt{\frac{4 n \log(1/\delta)}{\Delta^2 n_{\text{mix}}}} \) +  \sqrt{\frac{\Delta^2 n_{\text{mix}} \log(1/\delta)}{4 n }}. \label{Eq:DP}
\end{align}
The term $\Delta^2$ is given as $\Delta^2=d_x/\sigma_x^2 + d_y/\sigma_y^2$ where $d_x$ and $d_y$ are the sample and label dimensions (e.g., for the $28\times28$ pixel MNIST images of hand-written $0$-$9$ digits, $d_x=28 \times 28=784$ and $d_y=10$). As observed by $\varepsilon$ decreasing with $n$ in \eqref{Eq:DP}, Mixup can guarantee the raw sample privacy as long as the local dataset size is sufficiently large. Recall that this differential privacy analysis is based on $\gamma=0.5$ and additive noises. For more general cases under $\gamma>0$ without additive noise, we numerically evaluate the sample privacy by measuring the similarity between the raw and mixed-up samples in Chapt. \ref{Sec:Mix2FD_Simulation}.
}

\subsection{Proposed: Mix2FLD}
While MixFLD preserves local data privacy during seed sample collection, the Mixup operations may too significantly distorts the collected seed samples, which may hinder achieving high accuracy. To resolve this issue, Mix2FLD additionally applies the inverse-Mixup algorithm to MixFLD, thereby not only ensuring local data privacy but also achieving high accuracy. For the sake of clear explanation, we hereafter focus on a two-worker setting, where workers $c$ and $c'$ independently mix up the following two raw samples having symmetric labels:

\begin{itemize}
\addtolength{\itemindent}{15pt}
 \item Worker $c$: {$\mathbf{x}^c_i$} with {$\mathbf{y}^c_i= \{1,0\}$} and {$\mathbf{x}^c_{j}$} with {$\mathbf{y}^c_j=\{0,1\}$}
 \item Worker $c'$: {$\mathbf{x}^{c'}_{i'}$} with {$\mathbf{y}^{c'}_{i'}=\{0,1\}$} and {$\mathbf{x}^{c'}_{j'}$} with {$\mathbf{y}^{c'}_{j'}=\{1,0\}$},
\end{itemize}
where $\mathbf{y}^c_i$ is a one-hot encoded ground-truth label vector of $\mathbf{x}^c_i$, referred to as a \emph{hard label}. 
Following \eqref{Eq:mixup}, worker $c$ mixes up is local samples $\mathbf{x}^c_i$ and $\mathbf{x}^c_j$, yielding the mixed-up sample $\hat{\mathbf{x}}^c_{ij}$ corresponding to the mixed-up label $\{\gamma,1-\gamma\}$, referred to as its \emph{soft label}. Likewise, worker $c'$ superpositions $\mathbf{x}^{c'}_{i'}$ and $\mathbf{x}^{c'}_{j'}$, resulting in the mixed-up sample $\hat{\mathbf{x}}^{c'}_{i'j'}$ having the soft label $\{1-\gamma,\gamma\}$. The workers $c$ and $c'$ upload $\hat{\mathbf{x}}^c_{ij}$ and $\hat{\mathbf{x}}^{c'}_{i'j'}$ with their soft labels to the server.

Then, the server in Mix2FLD converts the soft labels back into hard labels, such that the converted samples contain more similar features of the hard-labeled real dataset, while being still different from the raw samples. To this end, the server applies the \emph{inverse-Mixup} that linearly combines $n_{s}$ mixed-up samples such that the resulting sample has a hard label. For the case of $C=10$ workers, as depicted in Figure \ref{fig_Mix2FLD}(b), with the above-mentioned symmetric setting, the server combines {$\hat{\mathbf{x}}^c_{ij}$} and {$\hat{\mathbf{x}}^{c'}_{i'j'}$}, such that the resulting {$\widetilde{\mathbf{x}}^{cc'}_{ij,i'j',l}$} has the $l$-th converted hard label as the ground-truth. This is described as:
  \begin{align}
    \widetilde{\mathbf{x}}^{cc'}_{ij,i'j',l}=\hat{\gamma}\hat{\mathbf{x}}^c_{ij}+(1-\hat{\gamma})\hat{\mathbf{x}}^{c'}_{i'j'}. \label{eq:invmixup}
    \end{align} 
The inverse mixing ratio $\hat{\gamma}$ for $n_{s} = 2$ is chosen in the following way. Suppose the target hard label is $\{1, 0\}$, i.e., $l=1$. Applying $\{1, 0\}$ to the LHS of \eqref{eq:invmixup} and $\{\gamma, 1-\gamma\}$ and $\{1-\gamma, \gamma\}$ of {$\hat{\mathbf{x}}^c_{ij}$} and {$\hat{\mathbf{x}}^{c'}_{i'j'}$} to the RHS of \eqref{eq:invmixup} yields two equations.\begin{align}
	1 &= \hat{\gamma} \gamma + (1-\hat{\gamma})(1-\gamma) \label{Eq:mix2up_1}\\
	0 &= \hat{\gamma}(1-\gamma) + (1-\hat{\gamma})\gamma \label{Eq:mix2up_2}
	\end{align} 
Solving these equations yields the desired $\hat{\gamma}$. By induction, this can be generalized to $n_{s} > 2$.

Hereafter, for the sake of convenience, we explain the rest of the algorithm considering $n_s = 2$. By alternating $\hat{\gamma}$ with $l=1$ and $2$, inversely mixing up two mixed-up samples {$\hat{\mathbf{x}}^c_{ij}$} and {$\hat{\mathbf{x}}^{c'}_{i'j'}$} yields two inversely mixed-up samples {$\widetilde{\mathbf{x}}^{cc'}_{ij,i'j',1}$} and {$\widetilde{\mathbf{x}}^{cc'}_{ij,i'j',2}$}. The server generates $n_{inv}$ inversely mixed-up samples by pairing two samples with symmetric labels among $n_{\text{mix}}$ mixed-up samples. By nature, inverse-Mixup is a data augmentation scheme, so $n_{inv}$ can be larger than $n_{\text{mix}}$. Note that none of the raw samples are identical to inversely mixed-up samples. To ensure this, inverse-Mixup is applied only for the seed samples uploaded from different devices, thereby preserving data privacy. The overall operation of Mix2FLD is summarized in Algorithm \ref{Algorithm_Mix2FLD}.

\begin{algorithm}[h]{
\small
		\caption{FLD with Mix2up (\textbf{Mix2FLD})}\label{Algorithm_Mix2FLD}
		\begin{algorithmic}[1]
			\Require $\mathcal{S}^c$ with $c\in \{1,\dots,C\}$, $\gamma\in(0,1)$
			\While  {not converged}
			
			
			\Procedure{Local Training and Mixup}{at worker $c\in \{1,\dots,C\}$}
			
			\State \textbf{if} $r = 1$ \textbf{generates} $\{\hat{\mathbf{x}}^c_{ij}\}$ via \eqref{Eq:mixup} \textbf{end if} \emph{\hfill$\triangleright$ Mixup}

			\State \textbf{updates} $\mathbf{w}^c$ and $\bar{F}^c_{l,r}$ for $K$ iterations as in FD (\textbf{Algorithm \ref{euclid}})
			
			\State \textbf{unicasts} $\{\bar{F}^c_{l,r}\}$ (with $\{\hat{\mathbf{x}}^c_{ij}\}$ \textbf{if} $r=1$) to the server
			
\EndProcedure			
			
			\Procedure  {Ensembling and Output-to-model conversion}{at server}
			
			\State \textbf{if} {$r=1$} \textbf{generates} $\{\widetilde{\mathbf{x}}^{cc'}_{ij,i'j',l}\}$ via \eqref{eq:invmixup} \textbf{end if} \emph{\hfill $\triangleright$ Inverse-Mixup}
			
			\State \textbf{computes} $\{\hat{F}^c_{l,r}\}$
			
			\State \textbf{updates} $\mathbf{w}_{g,k}$ via \eqref{eq:FLD} for $K_s$ iterations

			\State \textbf{broadcasts} $\mathbf{w}_{g,K_s}$ to all devices
			\EndProcedure
			\State $r\gets{r+1}$
			
			\State Worker $c\in\{1,\dots,C\}$ \textbf{substitutes} $\mathbf{w}^c_0$ with $\mathbf{w}_{g,K_s}$ \emph{\hfill$\triangleright$ Model download}
			\EndWhile{end while}    
			
		\end{algorithmic} }     
	
\end{algorithm}

\subsection{Numerical Evaluation and Discussions} \label{Sec:Mix2FD_Simulation}

\begin{figure*}[t!]
	\subfigure[Asymmetric channels, IID dataset.]{\includegraphics[width=0.495\columnwidth]{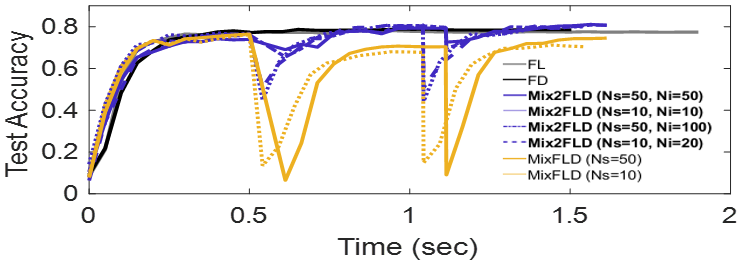}}
	\subfigure[Symmetric channels, IID dataset.]{\includegraphics[width=0.495\columnwidth]{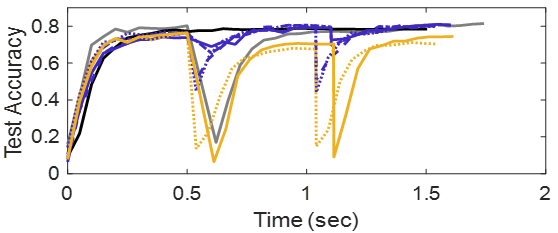}}
	\subfigure[Asymmetric channels, Non-IID dataset.]{\includegraphics[width=0.495\columnwidth]{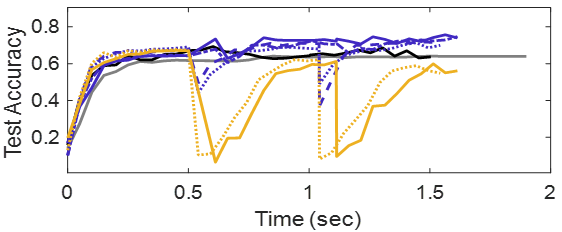}}
	\subfigure[Symmetric channels, Non-IID dataset.]{\includegraphics[width=0.495\columnwidth]{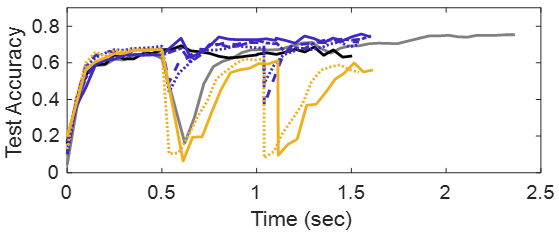}}
	\caption{all Learning curve of a randomly selected device in Mix2FLD, compared to FL, FD, and MixFLD, under asymmetric and symmetric channels, when $\gamma=0.1$ with IID and non-IID datasets.}\label{fig_Mix2FLDcompare}
\end{figure*}

In what follows, we provide a numerical performance evaluation of Mix2FLD compared with FL, FD, and MixFLD, in terms of the test accuracy and convergence time of a randomly selected reference device, under different data distributions (IID and non-IID) and uploaded/generated seed sample configurations: $(n_{\text{mix}}, n_{inv})\in\{(10,10),(10,20),(50,50),(50,100)\}$. The convergence time includes communication delays during the uplink and downlink, as well as the computing delays of devices and the server, measured using tic-toc elapsed time.

Every device has a $3$-layer convolutional NN model ($2$ convolutional layers, $1$ fully-connected layer) having $12,\!544$ model parameters in total. The server's global model follows the same architecture. Each worker owns its local MNIST dataset with $|\mathcal{Y}|=10$ classes and $n=500$ samples. For the IID case, every label has the same number of samples. For the non-IID case, randomly selected two labels have two samples respectively, while each of the other labels has $62$ samples. Other simulation parameters for model training are given as: $C=10$, $K=6,\!400$ iterations, $K_{s}=3,\!200$ iterations, $\eta=0.01$. The simulation parameters for reflecting wireless environment is given as the same as in \cite{MixFLD}.

The \emph{impact of channel conditions} is illustrated in Figure \ref{fig_Mix2FLDcompare}. The result shows that Mix2FLD achieves the highest accuracy with moderate convergence under asymmetric channel conditions among others. Compared to FL uploading model weights, Mix2FLD's model output uploading reduces the uplink payload size by up to $42.4$ times. Under asymmetric channels with the limited uplink capacity (Figures \ref{fig_Mix2FLDcompare}(a) and (c)), this enables more frequent and successful uploading, thereby achieving up to $16.7$\% higher accuracy and $1.2$ times faster convergence. Compared to FD, Mix2FLD leverages the high downlink capacity for downloading the global model weights, which often provides higher accuracy than downloading model outputs as reported in \cite{MLPCD}. In addition, the global information of Mix2FLD is constructed by collecting seed samples and reflecting the global data distribution, rather than by simply averaging local outputs as used in FD. Thereby, Mix2FLD achieves up to $17.3$\% higher accuracy while taking only $2.5$\% more convergence time than FD. Under symmetric channels, FL achieves the highest accuracy. Nevertheless, Mix2FLD still converges $1.9$ times faster than FL, thanks to its smaller uplink payload sizes and more frequent updates.

\begin{figure}[t!] 
	\begin{center}		
		\includegraphics[width=1.0\linewidth]{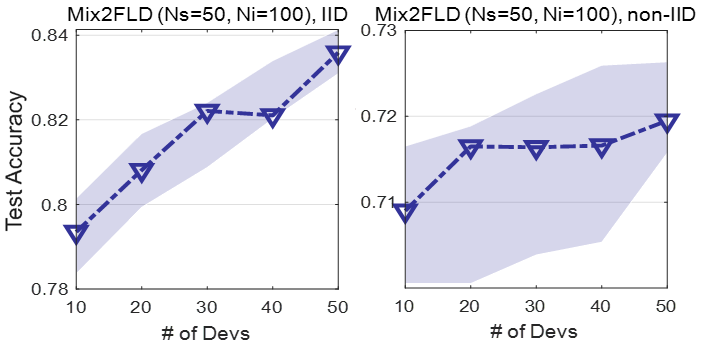}
	\end{center}
	\caption{ Test accuracy distribution of Mix2FLD w.r.t the number of devices, under symmetric channels with IID and non-IID datasets.}
	\label{Fig_Mix2FLD_scalability}
\end{figure}


Next, the \emph{impact of the number of devices} is observed in Figure \ref{Fig_Mix2FLD_scalability}. When the number of devices is increased from $10$ to $50$, the average of test accuracy increases by $5.7$\% and the variance decreases by $50$\% with IID dataset. In the non-IID dataset, the test accuracy gain is smaller than that of the IID dataset, but has the same tendency. This concludes that Mix2FLD is scalable under both IID and non-IID data distributions.

Furthermore, the \emph{effectiveness of Mix2up} is depicted in Figures \ref{fig_Mix2FLDcompare}(c) and (d), corroborating that Mix2FLD is particularly effective in coping with non-IID data. In our non-IID datasets, samples are unevenly distributed, and locally trained models become more biased, degrading accuracy compared to IID datasets in Figures \ref{fig_Mix2FLDcompare}(a) and (c). This accuracy loss can partly be restored by additional global training (i.e., output-to-model conversion) that reflects the entire dataset distribution using few seed samples. While preserving data privacy, MixFLD attempts to realize this idea. However, as observed in Figure \ref{fig_Mix2FLDcompare}(d), MixFLD fails to achieve high accuracy as its mixed-up samples inject too much noise into the global training process. Mix2FLD resolves this problem by utilizing inversely mixed up samples, reducing unnecessary noise. Thanks to its incorporating the data distribution, even under symmetric channels (Figure \ref{fig_Mix2FLDcompare}(d)), Mix2FLD achieves higher accuracy than FL. One drawback of Mix2up is its relying on an $n_s\times n_s$ matrix inversion for inverting $n_s$ linearly mixed-up samples, which may hinder the scalability of Mix2FLD for large $n_s$. Alternatively, as demonstrated in \cite{shin2020xor}, one can exploit the bit-wise XOR operation and its flipping property (e.g., $(A\oplus B)\oplus B = A$) replacing mixup and inverse-mixup, respectively, thereby avoiding the matrix inversion complexity.

\begin{table}[t!]
\footnotesize
	\centering
	\caption{Sample privacy, \emph{Mixup} ($n_{\text{mix}}$=100).}
	\resizebox{\columnwidth}{!}{\begin{tabular} {l c c c c c c}
			\toprule[1pt]
			\multirow{2}{*}{\textbf{Dataset}}& \multicolumn{6}{c}{\textbf{Sample Privacy} Under Mixing Ratio $\gamma$} \\
			&  $\gamma$ = 0.001 & 0.1 & 0.2 & 0.3 & 0.4 & 0.499   \\
			\midrule
			MNIST & 2.163 & 4.465 & 5.158 & 5.564 & 5.852 & \textbf{6.055} \\
			FMNIST& 1.825 & 4.127 & 4.821 & 5.226 & 5.514 & \textbf{5.717}  \\
			CIFAR-10& 2.582 & 4.884 & 5.577 & 5.983 & 6.270 & \textbf{6.473}  \\
			CIFAR-100& 2.442 & 4.744 & 5.438 & 5.843 & 6.131 & \textbf{6.334}  \\
			\bottomrule[1pt]
	\end{tabular}}   
\label{tab:Mixup}
\end{table} 

\begin{table}[t!]
\footnotesize
	\centering
	\caption{ Sample privacy, \emph{Mix2up} ($n_{\text{mix}}$=100).}
	\resizebox{\columnwidth}{!}{\begin{tabular} {l c c c c c c}
			\toprule[1pt]
			\multirow{2}{*}{\textbf{Dataset}}& \multicolumn{6}{c}{\textbf{Sample Privacy} Under Mixing Ratio $\gamma$} \\
			&  $\gamma$ = 0.001 & 0.1 & 0.2 & 0.3 & 0.4 & 0.499   \\
			\midrule
			MNIST & 2.557 & 4.639 & 5.469 & 6.140 & 7.007 & \textbf{9.366} \\
			FMNIST& 2.196 & 4.568 & 5.410 & 6.143 & 6.925 & \textbf{9.273}  \\
			CIFAR-10& 2.824 & 5.228 & 6.076 & 6.766 & 7.662 & \textbf{10.143}  \\
			CIFAR-100& 2.737 & 5.151 & 6.050 & 6.782 & 7.652 & \textbf{10.104}  \\
			\bottomrule[1pt]
	\end{tabular}}   
	\label{tab:Mix2up}
\end{table}


Lastly, the tradeoffs among latency, privacy, and accuracy are illustrated in Figure \ref{fig_Mix2FLDcompare}. For all the considered channel conditions and data distributions, in Mix2FLD and MixFLD, reducing the seed sample amount ($n_{\text{mix}}=10$) provides faster convergence time albeit compromising accuracy, leading to a \emph{latency-accuracy} tradeoff. The inverse-Mixup of Mix2FLD can partly resolve the tradeoff by more augmenting the seed samples. Even for the same $n_{\text{mix}}$, increasing $n_{inv}$ improves the accuracy by up to $1.7\%$. In doing so, the inverse-Mixup of Mix2FLD can increase the accuracy without additional communication latency.
Next, to validate the data privacy guarantees of Mixup and Mix2up, we evaluate the \emph{sample privacy}, given as the minimum similarity between a mixed-up sample and its raw sample: $\log (\min\{||\hat{\mathbf{x}}^c_{ij}-\mathbf{x}^c_{i}||,||\hat{\mathbf{x}}^c_{ij}-\mathbf{x}^c_{j}||\}\!)$ according to \cite{Jeong:FML19}. 
Table \ref{tab:Mixup} shows that Mixup ($\gamma>0$) with a single device preserves more sample privacy than the case without Mixup ($\gamma=0$). Table \ref{tab:Mix2up} illustrates that Mix2up with two devices preserves higher sample privacy than Mixup thanks to the additional (inversely) mixing up of the seed samples across devices. It also shows that each inversely mixed-up sample does not resemble its raw sample but an arbitrary sample having the same ground-truth label. Both Tables \ref{tab:Mixup} and~\ref{tab:Mix2up} show that the mixing ratio $\gamma$ closer to $0.5$ (i.e., equally mixing up two samples) ensures higher sample privacy, which may require compromising more accuracy. Investigating the \emph{privacy-accuracy} could be an interesting topic for future research.
  \section{Application: FD for Reinforcement Learning} \label{Chapt:FRD}

The original design of FD relies on grouping model outputs based on labels in classification. To demonstrate its applicability beyond classification, in this section we aim to exemplify an FD implementation under a reinforcement learning (RL) environment in which multiple interactive agents locally carry out decision-making in real time. In such environments, policy distillation (PD) is a well-known solution \cite{Rusu16}, wherein multiple agents collectively train their local NNs. As illustrated in Figure \ref{Fig:PD}, PD is operated by: (i) uploading every local \emph{experience memory} to a server, (ii) constructing a global experience memory at the server, and (iii) downloading and replaying the global experience memory at each agent to train its local NN \cite{Rusu16}. However, the local experience memory contains all local state observations and the corresponding policies (i.e., action logits). Exchanging such raw memories may thus violate the privacy of their host agents. Furthermore, the global experience memory size increases with the number of agents. The resulting ever-growing communication overhead may undermine the scalability of PD.

To obviate the aforementioned problems, by leveraging FD, we introduce \emph{federated reinforcement distillation (FRD)} \cite{FRD,Han:Intellisys20}, a communication-efficient and privacy-preserving distributed RL framework based on a \emph{proxy experience memory}. In FRD, each agent stores a local proxy experience memory that consists of a set of pre-arranged \emph{proxy states} and \emph{locally averaged policies}. In this memory structure, the actual states are mapped into the proxy states (e.g., based on the nearest value rule), and the actual policies are averaged over time. Exchanging the local proxy memories of agents not only preserves the privacy of agents, but also avoids the continuaal increase in the communication overhead as the number of agents grows. In what follows we first elaborate the baseline PD operations, and then illustrate FRD operations, followed by numerical evaluations.

\subsection{Policy Distillation With Experience Memory}
We consider an episodic environment modeled by a Markov decision process. The state space $\mathcal{S}$ and action space $\mathcal{A}$ are discrete. Without any prior knowledge on the environment, each agent takes an action $a\in\mathcal{A}$ at time slot $t$, and in return receives the reward $r_t \in \mathbb{R}$.
The resulting policy $\pi_\theta: \mathcal{S} \rightarrow \mathcal{P}(\mathcal{A})$, i.e., actions for given states, is stochastic, where $\mathcal{P}(\mathcal{A})$ is the set of probability measures on $\mathcal{A}$. The policy is described by the conditional probability $\pi_\theta(a|s)$ of $a\in\mathcal{A}$ for a given state $s\in\mathcal{S}$, where $\theta \in \mathbb{R}^n$ denotes the local model parameters of an agent. {Hereafter the subscript $c \in \{1,2, \cdots, C\}$ identifies an agent out of $C$ agents, and we abuse the notations by dropping it if the relationships are clear.}

\begin{figure}
	\centering
	\includegraphics[width=\linewidth]{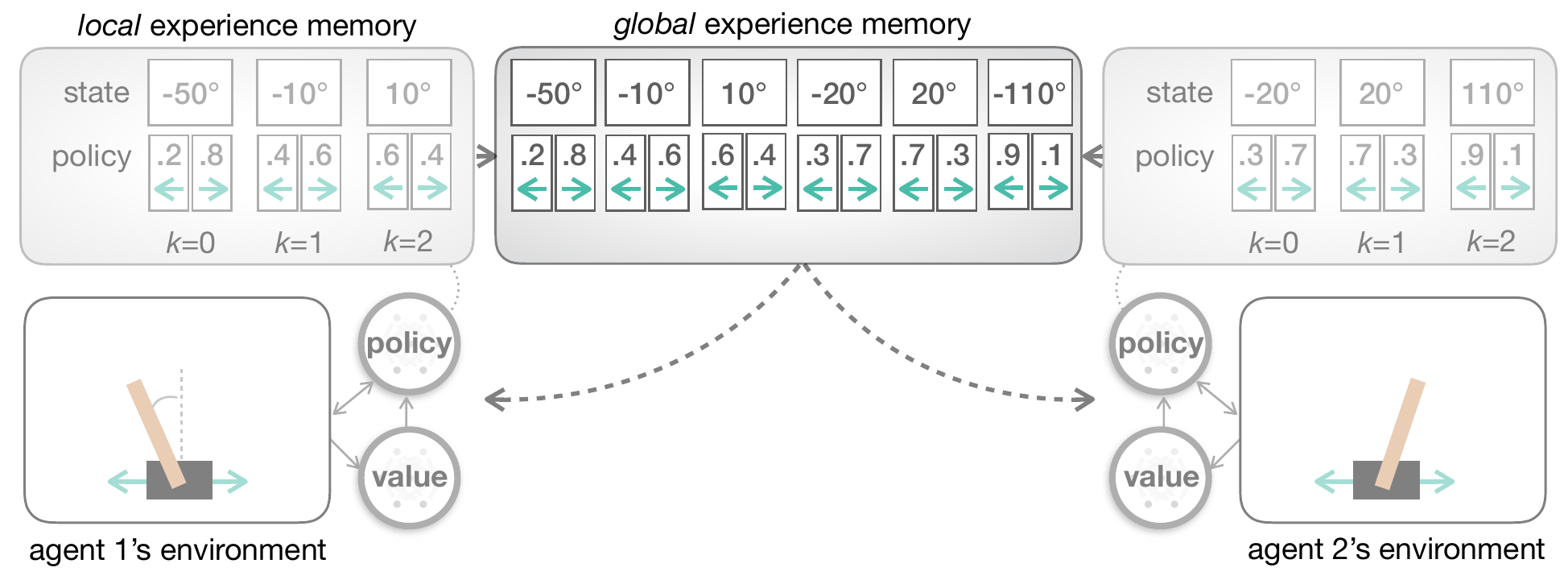}
	\caption{\small A schematic illustration of policy distillation (PD) with experience memory \cite{Rusu16}.} \label{Fig:PD}
	\end{figure}

{In PD \cite{Rusu16}, as depicted by Figure \ref{Fig:PD}, the agents collectively construct a dataset named \textit{experience memory} for training the local models.
The operation of PD can be summarized by the following steps.
\begin{enumerate}
\item Each agent records an \textit{local experience memory} $\mathcal{M}_c=\{(s_k, \pi_{\theta_c,k}(\mathbf{a}_k|s_k))\}^{K_c}_{k=1}$ for $E$ episodes. Note that $K_c$ is the size of local experience memory.
\item After all the agents complete $E$ episodes, the server collects the local experience memories from all agents.
\item The server constructs a \textit{global experience memory} $\mathcal{M}=\{(s_k, \pi_{\Theta,k}(\mathbf{a}_k|s_k))\}^{K}_{k=1}$, where $K=\sum_{c=1}^{C}K_c$ and $\pi_{\Theta,k}$ is the policy collected from the clients.
\item To reflect the knowledge of other agents, the agents download the global experience memory $\mathcal{M}$ from the server.
\item Similar to the conventional classification setting, the agent $c$ optimizes the local model $\theta_c$ by minimizing the cross entropy loss $L_c(\mathcal{M},\theta_c)$ between the policy of local model $\pi_{\theta_c}$ and the policy $ \pi_{\Theta} $ of global experience memories $\mathcal{M}$, where $L_c(\mathcal{M},\theta_c)$ is given as
\begin{align}
L_c(\mathcal{M}, \theta_c)=-\sum_{k=1}^{K}\pi_{\Theta,k}(\mathbf{a}_k|s_k)\log\left(\pi_{\theta_c}(\mathbf{a}_k|s_k)\right).
\end{align}
\end{enumerate}
}

Unfortunately, under the above-mentioned operations of PD, malicious agents and honest-but-curious server may sneak a look at all the previously visited states and taken actions of every agent, incurring privacy leakage issues. Furthermore the global experience memory size increases with the number of agents, limiting the scalability of PD,

\subsection{Federated Reinforcement~Distillation With Proxy Experience~Memory}

\begin{figure}
	\centering
	\includegraphics[width=\linewidth]{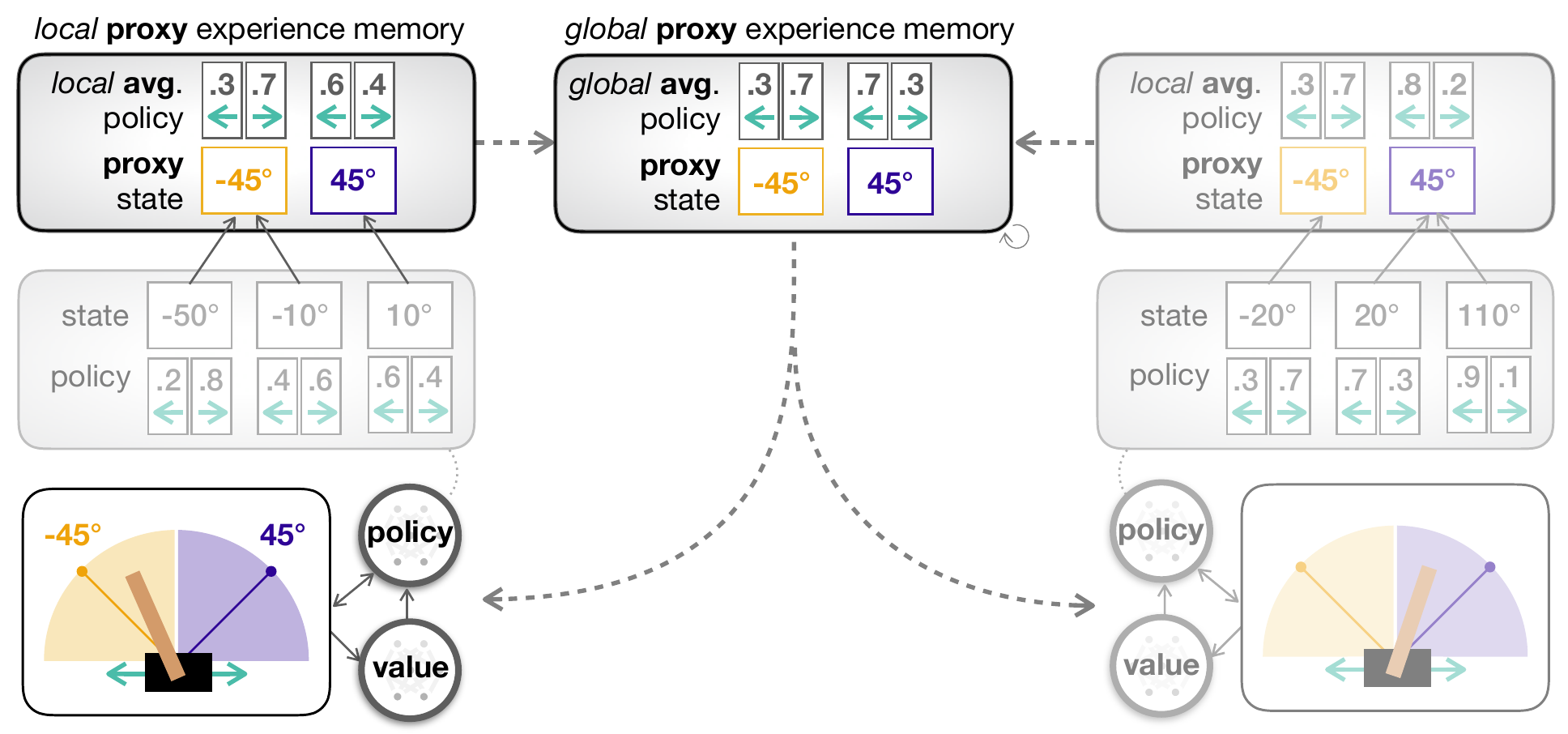}
	\caption{\small A schematic illustration of \textit{federated reinforcement distillation (FRD)} with \textit{proxy experience memory} \cite{FRD,Han:Intellisys20}.} \label{Fig:FRD}
\end{figure}

As opposed to PD, FRD relies on constructing and exchanging \textit{proxy experience memories} as illustrated in Figure \ref{Fig:FRD}, improving the communication efficiency while preserving privacy. {The proxy experience memory $\mathcal{M}^p$ is comprised of \textit{proxy state} $s^p$ and its associated \textit{average policy} $\pi_{\Theta}^p$. A proxy state is the representative state of each \textit{state cluster} $\mathcal{S}_j \subset \mathcal{S}$ for $i \in \{1,\dots,I\}$, where we assume $S_i \cap S_j = \emptyset$ for $i\neq j$. Given these definitions, the operations of FRD are described by the following steps. 
\begin{enumerate}
\item Each agent categorizes the experienced policy $\pi_{\theta_c,k}(\mathbf{a}|s)$ according to the proxy state cluster that the state $s$ is included in.
\item After all the agents complete the $E$ episodes, each agent constructs a \textit{local proxy experience memory} $\mathcal{M}^p_c=\{(s_{k'}^p,\pi_{\theta_i,k'}^p\left(\mathbf{a}_{k'}|s_{k'}^p\right)\}^{K_c^p}_{k'=1}$, where $ \pi_{\theta_c,k'}^p\left(\mathbf{a}_{k'}|s_{k'}^p\right)$ is the \textit{local average policy}, obtained by averaging the policy in the same category, while $K_c^p$ is the size of local proxy experience memory describing the number of proxy state clusters that have visited by the agent. Note that the $\pi_{\theta_i}^p\left(\mathbf{a}_k|s_k^p\right)$ is not generated by the local model of agent.
\item When the local proxy experience memory of every agent is ready, the server collects is from each agent.
\item Then, the server constructs the \textit{global proxy experience memory}
\begin{align}
\mathcal{M}^p=\{(s_{k'}, \pi_{\Theta,k'}(\mathbf{a}_{k'}|s_{k'}))\}^{K^p}_{k'=1}
\end{align}
by averaging the local average policies in the same category. The size of global proxy experience memory $K^p$ is the number of proxy state clusters that have visited by all the clients.
\item Each agent downloads the global proxy experience memory $ \mathcal{M}^p $ from the server.
\item Each agent $ i $ fits the local model $ \theta_c $ minimizing the cross entropy loss $ L^p_c(\mathcal{M}^p,\theta_c) $ between the policy of local model $\pi_{\theta_c}(\mathbf{a}_{k'}|s_{k'})$ and the global average policy $ \pi^p_{\Theta,k'}(s_{k'}^p,\mathbf{a}_{k'}|s_{k'}^p) $ of global proxy experience memory $ \mathcal{M}^p $, where
\begin{align}
L_c^p(\mathcal{M}^p, \theta_c)=-\sum_{k=1}^{K^p}\pi_{\Theta,k'}^p(\mathbf{a}_{k'}|s_{k'}^p)\log\left(\pi_{\theta_c,k'}(\mathbf{a}_{k'}|s_{k'}^p)\right).
\end{align}
This loss is calculated with the policy produced by the local model as the input of a proxy state.
\end{enumerate}
}


Constructing the local and proxy experience memories can be interpreted as quantizing the memories, thereby reducing the uplink and downlink payload sizes, respectively. Notably, the downlink payload size reduction significantly benefits from sharing each global proxy experience by multiple agents. This is in stark contrast to PD wherein the different agents' experiences are hardly overlapped with each other particularly for a large state dimension, bringing higher communication efficiency on FRD. Furthermore, exchanging proxy experience memories does not reveal any raw experiences of agents, enabling privacy-preserving distributed RL.

The local experiences are obtained by running a deep RL method at each agent. Throughout this chapter we consider the advantage actor-critic (A2C) framework~\cite{pmlr-v48-mniha16}, in which each agent stores a pair of actor and critic NNs. The actor NN generates an action $a \in \mathcal{A}$ according to the policy $\pi_\theta$, while the critic NN evaluates the benefit of the generated action compared to other possible actions, in terms of obtaining higher expected future reward. Since the actor and critic NNs have no prior knowledge on the environment, the actor-critic pair must interact with the environment, and thereby learn the optimal policy $\pi^*$ to gain the maximum expected future reward. Meanwhile, the benefit of taking an action is evaluated using the advantage function~$A^\pi(s_t,a_t)$ \cite{Wang:2016aa}, given as 
\begin{align} 
A^\pi(s_t,a_t) &= Q^\pi(s_t,a_t)-V^\pi(s_t) \\
&= r(s_t,a_t) + \mathbb{E}_{s_{t+1}\sim\mathbb{E}}\left[V^\pi(s_{t+1})\right]-V^\pi(s_t) \\
&\approx r(s_t,a_t) + V^\pi(s_{t+1}) - V^\pi(s_t),\label{eq:advantage}
\end{align}
where $ V^{\pi}\!(s)\!=\!\mathbb{E}[r_0^\gamma|s_0=s;\pi]$ is the value function, $Q^\pi\!(s,a)\!=\!\mathbb{E}\left[r_0^\gamma|s_0\!=\!s,a_0\!=\!a;\pi\right]$ is the Q-function, and $ r(s_t,a_t) $ is the instant reward at learning step $t$.
Note that if the output value of the advantage function is positive, it means that the selected action is not an optimal solution. Moreover, we can see from \eqref{eq:advantage} that the advantage function is approximately described only using the value function. The critic NN who computes the value can thereby evaluate the advantage for each updating step of the actor NN. The actor NN is a policy NN who approximates the policy $\pi$ and constructs the local experience memory. Lastly, in that each agent stores a pair of actor and critic NNs, there are three possibilities of exchanging: only actor NNs, critic NNs, or both actor and critic NNs across agents. As seen by several experiments \cite{FRD,Han:Intellisys20}, exchanging only actor NNs, i.e., policy NNs, achieves the convergence speed as fast as exchanging both actor and critic NNs, while saving the communication cost thanks to ignoring critic NNs. Hereafter we thus focus on an FRD implementation with the experience memory constructed by the actor NN outputs.

\subsection{Experiments and Discussions}
To show the effectiveness of FRD, we consider the \textit{CartPole-v1} environment in the OpenAI gym \cite{brockman2016openai}, where each agent controls a cart so as to make a pole attached to the cart upright as long as possible. Each agent obtains a score of $+1$ for every time slot during which the pole remains upright. Playing the CartPole game with multiple episodes, the agents complete a mission when any agent first reaches an average score of $490$, where the average is taken across $10$ latest episodes. 

The performance of FRD is evaluated in terms of the mission completion time, and is compared with two baseline distributed RL frameworks: PD \cite{Rusu16} and federated reinforcement learning (FRL) that exchanges actor NN model parameters following the standard FL operations \cite{McMahan2016,Sumudu:20,Kim:CL20,Google:FL19,Han:Intellisys20}. Each agent runs an A2C model comprising a pair of actor and critic NNs \cite{pmlr-v48-mniha16}, each of which is a multi-layer perceptron (MLP) with 2 hidden layers. At an interval of $25$ episodes, the agents exchanges their critic NN's outputs in PD and FRD or the critic NN parameters in FRL.

To construct proxy experience memories in FRD, the agent states are clustered as follows. In the \textit{Cartpole} environment, each agent has its 4-tuple state consisting of the cart location, cart velocity, pole angle, and the angular velocity of the pole. By evenly dividing each observation space into $S=30$ subspaces, we define state clusters as the combinations of the four subspaces, resulting in $S^4$ state clusters in total. A proxy state is defined by the middle value of each state cluster, and each raw state is mapped into the proxy state based on the nearest value rule. For example, the proxy state of the pole angle is $-45^\circ$ when the state cluster is $[-90^\circ, 0^\circ)$, as illustrated in Figure \ref{Fig:FRD}. Throughout the simulations, the lines represent the median values, and the shaded areas depict the regions between the top-25 and top-75 percentiles.

\begin{figure}
	\centering
	\subfigure[Mission completion time.]{\includegraphics[width=.6\linewidth]{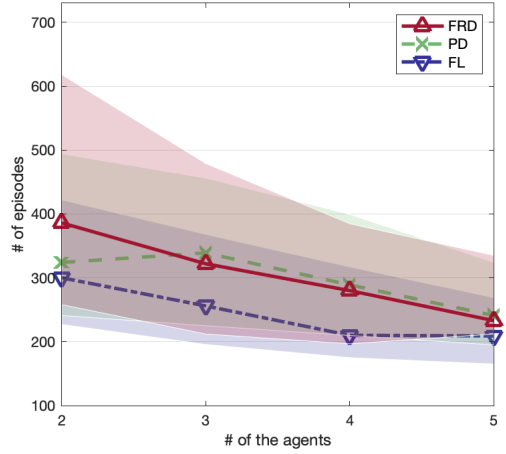}}
	\subfigure[Communication cost.]{\includegraphics[width=.6\linewidth]{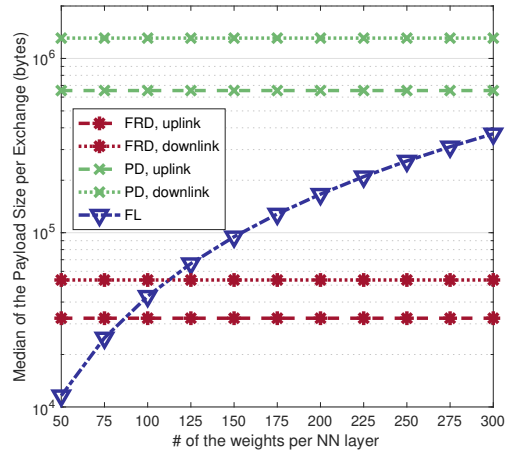}}
	\caption{Performance of FRD compared to PD and FRL, in terms of (a) the mission completion time and (b) communication cost.
	} \label{Fig:FRD_sim}
	\end{figure}

	In comparison with PD, FRD achieves the mission completion time as fast as PD as shown by Figure~\ref{Fig:FRD_sim}(a), while saving the communication cost by around 50\% as observed by Figure~\ref{Fig:FRD_sim}(b) for $2$ agents. In Figure~\ref{Fig:FRD_sim}(b), the payload size gap between the uplink and downlink is due to the difference between local and global (proxy) experience memory sizes. This uplink-downlink payload size gap of PD is larger than that of FRD for $2$ agents, which is expected to become even larger for more agents thanks to the proxy state sharing of FRD, advocating the communication efficiency and scalability of FRD.

	Compared to FRL, FRD completes the mission slightly slower than FRL particularly for a small number of agents, as illustrated in Figure~\ref{Fig:FRD_sim}(a). However, the the communication payload size of FRL increases with the actor NN model size, incurring higher payload sizes than FRD when there are over $100$ neurons per layer as depicted by Figure~\ref{Fig:CD_sim}(b). Furthermore, due to the nature of exchanging and averaging model parameters, all the agents under FRL are forced to have an identical critic NN architecture, limiting the adoption of FRL particularly for a large-scale implementation with heterogeneous agents. By contrast, FRD yields the communication cost upper bounded by the number of state clusters, and does not impose any constraint on the NN architecture selection, highlighting the communication efficiency and flexibility of FRD.

  \section{Conclusion} \label{Chapt:Conclusion} 

In this chapter we introduced federated distillation (FD), a distributed learning framework that exchanges model outputs as opposed to federated learning (FL) based on exchanging model parameters. FD leverages key principles of co-distillation (CD), an online version of knowledge distillation (KD), and pushes the frontiers of its communication efficiency forward via a novel model output grouping method. To provide a deep understanding of FD, we provided a neural tangent kernel (NTK) analysis of CD in a classification task, proving that CD asymptotically achieves the convergence to the ground-truth prediction even with two workers, while more workers accelerate the convergence speed. Treating CD as the method providing the upper bound accuracy of FD, while still effective in terms of communication efficiency, our vanilla implementation of FD is far from achieving the maximum achievable accuracy. To fill this gap, we presented several advanced FD applications harnessing wireless channel characteristics and/or exploiting proxy datasets, thereby achieving even higher accuracy than FL. The potential of FD is not limited to classification tasks. We partly advocated such possibilities of FD by exemplifying a reinforcement learning (RL) use case. Going beyond this, for future research, it could be worth studying the applicability of FD to unsupervised learning and self-supervised learning tasks under more realistic wireless channels and time-varying network topologies.

\bibliographystyle{IEEEtran}
\bibliography{FDbook}


\end{document}